\documentclass{article}
\usepackage{CJKutf8}

\usepackage{float}

\usepackage[sort,numbers]{natbib}
\setcitestyle{square}

\usepackage{amssymb}

\usepackage[preprint]{neurips_2024}

\usepackage{amsmath}

\DeclareMathOperator*{\argmin}{arg\,min}

\usepackage{adjustbox}
\usepackage{array}
\usepackage{diagbox}

\usepackage{enumitem}

\makeatletter
\newcommand{\rmnum}[1]{\romannumeral #1}
\newcommand{\Rmnum}[1]{\expandafter\@slowromancap\romannumeral #1@}
\makeatother

\usepackage[utf8]{inputenc} 
\usepackage[T1]{fontenc}    
\usepackage{hyperref}       
\usepackage{url}            
\usepackage{booktabs}       
\usepackage{amsfonts}       
\usepackage{nicefrac}       
\usepackage{microtype}      
\usepackage{xcolor}         
\usepackage{amsthm}
\usepackage{amsmath}

\usepackage[linesnumbered,ruled]{algorithm2e}

\usepackage{graphicx}
\usepackage{epstopdf}
\usepackage{natbib}
\usepackage{booktabs}
\usepackage{multirow}
\usepackage{graphicx}
\usepackage{booktabs}
\usepackage{graphicx}
\usepackage{subcaption}
\title{Benchmarking Data Heterogeneity Evaluation Approaches for Personalized Federated Learning}

\author{%
  Zhilong Li$^{1}$, Xiaohu Wu$^{1}$, Xiaoli Tang$^{2}$, Tiantian He$^{3,4}$, Yew-Soon Ong$^{2,3,4}$ \\
  \textbf{Mengmeng Chen$^{1}$, Qiqi Liu$^{5}$, Qicheng Lao$^{1}$, Han Yu$^{4}$}\\
$^{1}$Beijing University of Posts and Telecommunications, China \\
$^{2}$College of Computing and Data Science, Nanyang Technological University, Singapore\\
$^{3}$Institute of High Performance Computing, Agency for Science, Technology and Research, Singapore\\
$^{4}$Centre for Frontier AI Research, Agency for Science, Technology and Research, Singapore\\
$^{5}$School of Engineering, Westlake University, China\\
 \texttt{\{lzl,xiaohu.wu,chenmengmeng0306,qicheng.lao\}@bupt.edu.cn}\\ 
 \texttt{\{xiaoli001,asysong,han.yu\}@ntu.edu.sg}, he\_tiantian@cfar.a-star.edu.sg  \\
 \texttt{liuqiqi@westlake.edu.cn}  \\
}

\begin{document}

\maketitle

\begin{abstract}
There is growing research interest in measuring the statistical heterogeneity of clients' local datasets. Such measurements are used to estimate the suitability for collaborative training of personalized federated learning (PFL) models. Currently, these research endeavors are taking place in silos and there is a lack of a unified benchmark to provide a fair and convenient comparison among various approaches in common settings. We aim to bridge this important gap in this paper. The proposed benchmarking framework currently includes six representative approaches. Extensive experiments have been conducted to compare these approaches under five standard non-IID FL settings, providing much needed insights into which approaches are advantageous under which settings. The proposed framework offers useful guidance on the suitability of various data divergence measures in FL systems. It is beneficial for keeping related research activities on the right track in terms of: (\rmnum{1}) designing PFL schemes, (\rmnum{2}) selecting appropriate data heterogeneity evaluation approaches for specific FL application scenarios, and (\rmnum{3}) addressing fairness issues in collaborative model training. The code is available at \href{https://github.com/Xiaoni-61/DH-Benchmark}{https://github.com/Xiaoni-61/DH-Benchmark}.
\end{abstract}

\section{Introduction}


Federated learning (FL) is a promising privacy-preserving collaborative machine learning (ML) paradigm \cite{Yang19a,KairouzYu21a}. Under FL, multiple FL clients train a shared model with their own datasets and upload their local model updates to a FL server for aggregation and redistribution \cite{McMahan17a}. One key challenge of FL is that the data distributions across different clients can be heterogeneous (i.e., non-independently and identically distributed (non-IID)) \cite{zhu2021federated,NonIID-Benchmarking-ICDE-22,li2020fedbn}. 
This has inspired the field of personalized federated learning (PFL) \cite{Tan23a} to build more powerful ML models, better realizing the final goal of FL.

The non-IIDness of clients' local data entails evaluating data complementarity or similarity between clients and building a personalized ML model for each client. 
A FL system itself is a collaborative network of clients where a client $i$ may be complemented by other clients $j$ with different weights, which measure the data heterogeneity. The basic way that FL works is to aggregate the clients’ local model updates according to their weights, e.g., the vanilla Federated Averaging framework \citep{mcmahan2017communication}. In the context of PFL, we believe that it is still of fundamental importance to obtain a personalized model for each client by aggregating the clients' local model updates according to the weights.
Measuring statistical heterogeneity of data is one way to understand the data complementarity and potential collaboration advantages among clients \citep{Measuring-Data-Similarity-24a}. Recently, there has been growing interest in measuring data heterogeneity. These techniques are either divergence-based measures (e.g., Jensen–Shannon (JS) divergence \cite{Ding-NIPS-22,lu2024leap}, $\mathcal{C}$-divergence \cite{FedCollab-ICML-23,Long-23-WWW}, hash function-based distribution sketches \cite{DSketch-NIPS-23}), or model-based approaches (e.g., Shapley Value \cite{Shapley-TMC-24,GTG-Shapley-TIST-22}, Hypernetworks \cite{navon2020learning,Hypernetwork-21a}, cosine similarity \cite{Cosine-Similarity-ICML-23}). 
Although important, current research endeavors take place in silos and some are even not specifically designed for PFL. A comprehensive study is needed to simultaneously compare all the schemes in common settings of PFL. 

To this end, we develop a comprehensive benchmark to study the various approaches mentioned above. The main contributions of this paper are as follows. 
\begin{enumerate}[leftmargin=.16in]
    \item We summarize the six techniques (JS divergence, $\mathcal{C}$-divergence, distribution sketch-based euclidean distance, Shapley Value, Hypernetworks, cosine similarity) into a unified framework to understand their application in FL settings. The unified framework clarifies the ways of quantifying the collaboration advantages among clients and the theoretical development of using collaboration advantages or data similarity for PFL. 
    \item We evaluate all six approaches under five standard Non-IID settings summarized in \cite{NonIID-Benchmarking-ICDE-22} across
eight widely-adopted benchmark datasets. We assess the performance of each approach in terms of
computation cost, communication overhead and scalability. 
The results provide insights into which approaches are advantageous under which settings. 
    \item The unified framework and the experimental results identify scenarios where the current approaches perform relatively poorly, highlighting promising future research directions for collaborative PFL.
\end{enumerate}

\section{Preliminary}
\subsection{FL Notation}
\label{sec.FL-notation}
 
Consider a FL setting with $N$ clients. 
Each client $i\in \{1, 2, \dots, N\}$ has a dataset $\hat{D}_{i}=\{ (x_{k}^{(i)},\, y_{k}^{(i)})\}_{k=1}^{m_{i}}$ consisting of $m_{i}$ samples drawn from an underlying data distribution $\mathcal{D}_{i}$, where $x_{k}^{(i)}\in \mathcal{X}$ denotes the feature while $y_{k}^{(i)}\in \mathcal{Y}$ denotes the label. The total number of samples across all clients is $m=\sum_{i=1}^{N}{m_{i}}$. Let $\beta_{i}=m_{i}/m$ denote the proportion of data samples held by client $i$, and $\beta=[\beta_{1}, \dots, \beta_{N}]$ represent the quantity distribution of data samples across clients. Given a machine learning model (hypothesis) $h$ and a risk function $\boldsymbol\ell$, the local expected risk of client $i$ is defined as $\mathcal{L}_{i}(h) = \mathbb{E}_{(x,y)\in \mathcal{D}_{i}}{\boldsymbol\ell(h(x),y)}$ 
and its local empirical risk is given by $\hat{\mathcal{L}}_{i}(h) = \frac{1}{m_{i}}\sum\nolimits_{k=1}^{m_{i}}{\boldsymbol\ell(h(x_{k}^{(i)}),y_{k}^{(i)})}$. 
The goal of each client $i$ is to find a model $h$ within the hypothesis space $\mathcal{H}$ that minimizes its local expected risk, denoted as $h_{i}^{\ast}=$ $\argmin_{h\in\mathcal{H}}{\mathcal{L}_{i}(h)}$,
based on its finite local dataset $\hat{D}_{i}$.

We denote $w_i^t$ as the model parameters of client $i$ at the beginning of the training round $t$. Let $\mathbf{\alpha}_{i}^{t}=(\alpha_{i,1}^{t}, \dots \alpha_{i,N}^{t})$ denote the weighted collaboration vector of client $i$ in the training round $t$, where $\sum_{j=1}^{N}{\alpha_{i,j}^{t}}=1$ and $\alpha_{i,j}^{t}\geq 0$. 
Except the scheme in \citep{DSketch-NIPS-23}, the central server performs a weighted aggregation of the model parameters from all clients for each client $i$.
As a result, the model parameters of $i$ become $w_{j}^{t+1}=\sum_{j=1}^{N}{\alpha_{i,j}^{t}w_{j}^{t}}$. The weight $\alpha_{i,j}^{t}$ can be viewed as a quantification of the collaboration advantage that client $j$ brings to client $i$ in the training round $t$. Let $\alpha^{t}$ be a $N\times N$ matrix whose $i$-th row is $\alpha_{i}^{t}$. $\alpha^{t}$ defines a directed benefit graph among the $N$ clients. In the schemes of \citep{Cosine-Similarity-ICML-23,Shapley-TMC-24}, the value of $\alpha_{i,j}^{t}$ changes in each round of training. In the schemes of \citep{cui2022collaboration,Ding-NIPS-22,FedCollab-ICML-23}, the value of $\alpha_{i,j}^{t}$ is independent of the round $t$ and is precomputed before the FL training process starts; thus, $\alpha_{i,j}^{t}$, $\alpha_{i}^{t}$ and $\alpha^{t}$ are also simply denoted as $\alpha_{i,j}$, $\alpha_{i}$ and $\alpha$. The way that the scheme in \citep{DSketch-NIPS-23} works will be introduced in Section \ref{sec.RACE}.


\subsection{Typical Non-IID Data Settings}

From a probability distribution perspective, the local data distribution $P(x_{i}, y_{i})$ of client $i$ can be represented by a conditional probability $P(x_{i}|y_{i})P(y_{i})$ or $P(y_{i}|x_{i})P(x_{i})$. There are three typical non-IID data settings in the FL context \citep{Tan23a,NonIID-Benchmarking-ICDE-22,KairouzYu21a}: (\rmnum{1}) {\em label distribution skew}, where $P(y_{i})$ differs among clients; (\rmnum{2}) {\em feature distribution skew}, where $P(x_{i})$ differs among clients; and (\rmnum{3}) {\em quantity skew}, where $P(x_{i}, y_{i})$ is the same for all clients $i$, but the amount of data varies across clients.

The label distribution skew can be categorized into two settings:
\begin{itemize}[leftmargin=.16in]
    \item {\em Quantity-based label imbalance}: Each client is randomly allocated $k$ different label IDs \citep{McMahan17a,li2020federated,geyer2017differentially,yu2020federated}. For each label,  the samples are randomly and equally divided among the clients assigned to that label. We denote this setting as $\#C = k$. 
    \item {\em Distribution-based label imbalance}: A Dirichlet distribution $Dir_{N}(\epsilon)$ is used \citep{yurochkin2019bayesian,wang2020tackling,wang2019federated}. For each label $c$, a vector $q_{c}\in \mathbb{R}^{N}$ is drawn from the distribution $Dir_{N}(\epsilon)$ and client $i$ is allocated a $q_{c,i}$ proportion of the data samples for label $c$. We denote this setting as $p_{k}\sim Dir(\epsilon)$. 
\end{itemize}

The feature distribution skew mainly has two settings:
\begin{itemize}[leftmargin=.16in]
    \item {\em Noise-based feature imbalance}: The entire dataset is randomly and equally divided among $N$ clients \citep{zhang2017beyond}. Given a noise level $\sigma$, noises $\boldsymbol{\hat{x}} \sim Gau(\sigma\cdot i/n)$ are added to the (local) dataset of each client $i$, where $Gau(\sigma\cdot i/n)$ is a Gaussian distribution with mean 0 and variance $\sigma\cdot i/n$. We denote this setting as $\mathbf{\hat{x}}\sim Gau(\sigma)$.
    \item {\em Real-world feature imbalance}: The entire dataset is randomly and evenly partitioned among different clients \citep{caldas2018leaf}, with each client receiving data sharing the same real-world characteristics, such as being from the same writers.
\end{itemize}

 
In the {\em quantity skew} setting, a vector $q\in \mathbb{R}^{N}$ is drawn from the distribution $Dir_{N}(\epsilon)$, and each client $i$ is allocated a $q_{i}$ proportion of the total data samples of a dataset. We denote this setting as $q\sim Dir(\epsilon)$.



\section{Data Heterogeneity Evaluation Benchmark}
\label{sec.pFL-framework}
Existing methods for evaluating data heterogeneity can be roughly grouped into two main categories: those based on statistical divergence of data distributions and those based on model performance. In the following sections, we will detail these two categories of methods and their applications in PFL. 

\subsection{Distribution Statistical Divergence-based Approaches}
Methods in this category are based on the divergence,  a kind of statistical distance measure. It is a function that takes two probability distributions as input and returns a numerical value quantifying the difference between them \citep{rezende2018short,Measuring-Data-Similarity-24a}. In FL, the local model updates are uploaded to the central server, which performs a weighted aggregation of these updates to produce either a global model or a personalized models for each client. The divergence of data distributions among clients can guide more informed decisions on the choice of the aggregation weights \citep{Ding-NIPS-22,lu2024leap,FedCollab-ICML-23} or clustering clients with similar data distributions \citep{FedCollab-ICML-23,DSketch-NIPS-23,Long-23-WWW}.



Some works in this category \citep{Ding-NIPS-22,FedCollab-ICML-23} have some features in common. 
In these methods, the weighted empirical risk $\hat{\mathcal{L}}_{\boldsymbol{\alpha_{i}}}(h)$ for client $i$ is defined as $\hat{\mathcal{L}}_{\boldsymbol{\alpha_{i}}}(h)=\sum_{j=1}^{N}{\alpha_{i,j}\hat{\mathcal{L}}_{j}(h)}$. The model of $i$ can be learned by minimizing $\hat{\mathcal{L}}_{\boldsymbol{\alpha_{i}}}(h)$. Let $\hat{h}_{\boldsymbol{\alpha_{i}}}=\argmin_{h\in\mathcal{H}}{\hat{\mathcal{L}}_{\boldsymbol{\alpha_{i}}}(h)}$ be the hypothesis that minimizes the weighted empirical risk $\hat{\mathcal{L}}_{\boldsymbol{\alpha_{i}}}(h)$. 
The following bound is used by both \citep{Ding-NIPS-22} and \citep{FedCollab-ICML-23}: 
\begin{align}\label{equa-nju}
\mathcal{L}_{i}(\hat{h}_{\boldsymbol{\alpha_{i}}}) - \mathcal{L}_{i}(\hat{h}_{i}^{\ast}) \leqslant Q_{\gamma,1}\sqrt{\sum\nolimits_{j=1}^{N}{\alpha_{i,j}^2/m_j}} + Q_{\gamma,2}\sum\nolimits_{j=1}^{N} \alpha_{i,j}\cdot D(\mathcal{D}_i, \mathcal{D}_j)
\end{align}
where $Q_{\gamma,1}$ and $Q_{\gamma,2}$ are parameters related to the type of divergence used in \citep{Ding-NIPS-22,FedCollab-ICML-23}. Eq. (\ref{equa-nju}) shows that for client $i$ the gap between $\hat{h}_{\boldsymbol{\alpha_{i}}}$ and $\hat{h}^{\ast}$ depends on the weight $\alpha_{i}$, the number of samples $m_{i}$ and the divergence $D(\mathcal{D}_i, \mathcal{D}_j)$ between two data distributions. The optimal model for $i$ can be obtained by choosing appropriate decision variables to minimize the right-hand side of Eq. (\ref{equa-nju}). These methods are based on two statistical divergences, Jensen–Shannon Divergence and $\mathcal{C}$-Divergence.


\subsubsection{Jensen–Shannon Divergence}

The Jensen-Shannon (JS) divergence between two distributions $P$ and $Q$ is denoted as $\text{JSD}(P, Q)$ \citep{AIP04a}. It is based on the Kullback-Leibler (KL) divergence, denoted as $\text{KLD}(P, Q)$. For two discrete probability distributions on the same space $\mathcal{Z}$, we have $\text{KLD}(P, Q)=\sum\nolimits_{z\in\mathcal{Z}}{P(z)\log{\frac{P(z)}{Q(z)}}}$ and $\text{JSD}(P, Q) = \frac{1}{2} \text{KLD}(P, M) + \frac{1}{2} \text{KLD}(Q, M)$ where $M=\frac{1}{2}(P+Q)$ denotes the mixture distribution of $P$ and $Q$. The JS divergence has desirable properties, such as being symmetric and always having a finite value.

\vspace{0.1em}\noindent\textbf{Application.} \citet{Ding-NIPS-22} use the JS divergence to estimate the Integral Probability Metrics (IPM), which are another type of statistical distance  with desirable theoretical properties for analytical tractability. The IPM between two distributions \(\mathcal{D}_i\) and \(\mathcal{D}_j\) are denoted as \(d_{\mathcal{H}}(\mathcal{D}_i, \mathcal{D}_j)\), which is used by \citet{Ding-NIPS-22} to instantiate the divergence term $D(\mathcal{D}_{i}, \mathcal{D}_{j})$ on the right-hand side of Eq. (\ref{equa-nju}). Furthermore, \citet{Ding-NIPS-22} provides an optimal solution  $\alpha_{i}$ to the corresponding optimal problem. The resulting PFL scheme is called \textbf{pFedJS}. 


\subsubsection{$\mathcal{C}$-Divergence} 

\(\mathcal{C}\)-divergence is defined as \citep{mohri2012new,wu2020continuous,ben2010theory}: $D_{c}(\mathcal{D}_i, \mathcal{D}_j) = \max_{h \in \mathcal{H}} |\mathcal{L}_i(h) - \mathcal{L}_j(h)|$. In the hypothesis space $\mathcal{H}$, $D_{c}(\mathcal{D}_i, \mathcal{D}_j)$ indicates the maximum difference between the local expected risks of two data distributions $\mathcal{D}_i$ and $\mathcal{D}_j$. Suppose \(\ell\) is the 0-1 loss function. The definition of \(\mathcal{C}\)-divergence guarantees that its divergence value will be as close to one as possible if two distributions are distinctly different. Thus, it can be used to cluster clients into groups with similar data distributions. 

\vspace{0.1em}\noindent\textbf{Application.} In the FL context, $\mathcal{C}$-divergence has been used to partition all clients into $K$ coalitions $\{\mathcal{C}_{1}, \cdots, \mathcal{C}_{k}\}$, each with similar data distributions \citep{FedCollab-ICML-23,Long-23-WWW}.  The function $\ell(m(x), y)$ is also denoted as \(f(x, y)\), mapping from \(\mathcal{X} \times \mathcal{Y}\) to \(\{0, 1\}\). \citet{FedCollab-ICML-23} transform the $\mathcal{C}$-divergence into the following form: $D_{i,j} = \max\nolimits_{f \in \mathcal{F}} \left| \Pr\nolimits_{(x,y) \in D_i} [f(x, y) = 1] + \Pr\nolimits_{(x,y) \in D_j} [f(x, y) = 0] - 1 \right|$. 
A client classifier $f\in\mathcal{F}$ can be trained to predict the distribution similarity. If the estimated distance $D_{i,j}$ is approximately $ 100\%$, then the two distributions being compared are significantly different. 
Unlike in \citep{Ding-NIPS-22}, $\alpha_{i}$ is a dependent variable here, and the weight $\alpha_{i,j}$ of $j$ to $i$ is predefined as the proportion of the data quantity of $j$ to the coalition's data quantity. Specifically, for any $i\in\mathcal{C}_{k}$, $\alpha_{i,j}=\beta_{j}/\sum_{l\in\mathcal{C}_{k}}{\beta_{l}}$ if $j\in\mathcal{C}_{k}$ and $\alpha_{i,j}=0$ otherwise. Here, $D_{c}(\mathcal{D}_i, \mathcal{D}_j)$ is used to instantiate the divergence term $D(\mathcal{D}_{i}, \mathcal{D}_{j})$ on the right-hand side of Eq. (\ref{equa-nju}). \citet{FedCollab-ICML-23} provides an efficient optimizer to optimize the coalition structure. Once $\{\mathcal{C}_{1}, \cdots, \mathcal{C}_{k}\}$ is determined, $\{\mathbf{\alpha}_{i}\}_{i=1}^{N}$ will be determined accordingly. 
The corresponding PFL scheme is called \textbf{FedCollab}.




\subsubsection{
Distribution Sketch-based Euclidean Distance}
\label{sec.RACE}

Let $\mathbf{z_{i}}=(z_{i,1}, \dots, z_{i,d})$, where $i\in \{1,2\}$. The Euclidean distance of two vectors is defined as $\sqrt{\sum_{l=1}^{d}{(z_{1,l}-z_{2,l})^{2}}}$. On the other hand, efficient computation of statistical divergences of high-dimensional data is challenging. To address this, \citet{DSketch-NIPS-23} proposes a one-pass distribution sketch algorithm to represent the client data distributions. Then, the divergence is measured by the Euclidean distance of such distribution sketches. Central to their algorithm is Locality Sensitive Hashing (LSH), which uses hash functions to map data samples from high-dimensional space to low-dimensional buckets, ensuring that samples close to each other have a high probability of having the same hash value. Building on this, the Repeated Array of Count Estimators (RACE) is adopted \citep{coleman2020sub,coleman2022one,Array-18a,liu2022retaining}. Let $z=(x,y)$ denote a data sample. RACE uses \( R \) LSH functions to construct a matrix \( \boldsymbol{A} \in \mathbb{R}^{R \times B} \). Each LSH function \( g_i \) maps a data vector into \( B \) hash bins. For a given dataset \( \hat{\mathcal{D}} \), any data sample \( z \in \hat{\mathcal{D}} \) is mapped to \( R \) different hash values, denoted as \(\{ g_i(z) \mid i \in [1, R] \}\). The corresponding positions in matrix \( A \) are then incremented by 1. In this way, RACE can characterize the data distribution of \( \hat{\mathcal{D}}_{i} \) for each client $i$ using the corresponding \( A \), referred to as the client data sketch \( CS_{i} \).

\vspace{0.1em}\noindent\textbf{Application.} \citet{DSketch-NIPS-23} show that an average of all client sketches can effectively approximate the sketch  \( GS \) of the global data distribution across clients. The central server calculates the Euclidean distance between \( GS \) and $CS_i$ before the FL training process starts. The inverses of these distances serve as the probabilities for selecting \( K \) clients, denoted as $\mathcal{C}$ for local training in round \( t \). Specifically, clients with $CS_{i}$ closer to \( GS \) are prioritized for collaboration. The local updates from $\mathcal{C}$ are aggregated by the central server, resulting in a global model $w^{t+1}=\sum_{j\in\mathcal{C}}{w_{j}^{t}}$. Finally, each client builds a personalized model based on the global model and its local data. We refer to the PFL approach in \citep{DSketch-NIPS-23} as \textbf{RACE}.

\subsection{Model-based Approaches}

\subsubsection{Shapley Value}

Shapley Value (SV) is a classic approach to quantifying individual contributions within a group under the Cooperative Game Theory \citep{shapley1953value}. 
Let $\mathcal{N}=\{1, \dots, N\}$. In machine learning, the data Shapley value of client $i$ is defined as $\varphi_{i}(\mathcal{N}, V)=\sum\nolimits_{\mathcal{S}\subseteq \mathcal{N}-\{i\}}{\frac{V(\mathcal{S}\cup\{i\})-V(\mathcal{S})}{\binom{N-1}{|\mathcal{S}|}}}$, where $V(\cdot)$ denotes the utility evaluation function \citep{ghorbani2019data,GTG-Shapley-TIST-22,huang2022efficient}. In FL, the utility evaluation function $V(\mathcal{S})$ is based on the model performance achieved by the participation of $\mathcal{S}$. Let $M_{\mathcal{S}}$ denote the global FL model trained with $\mathcal{S}$ on a separate test dataset $\mathcal{D}_{\mathcal{S}}$. Then, $V(\mathcal{S})=V(M_{\mathcal{S}})=V\left(\mathcal{A}(M^{(0)}, \mathcal{D}_{\mathcal{S}})\right)$, where $\mathcal{A}$ is the learning algorithm and $M^{(0)}$ denotes the initial model \citep{GTG-Shapley-TIST-22}. 

\vspace{0.1em}\noindent\textbf{Application.} \citep{Shapley-TMC-24} applies the SV technique for PFL. For each client $i$, there is an $N$-dimensional relevance vector $\phi^{i,t}=[\phi_{1}^{i,t}, \dots ,\phi_{N}^{i,t}]$, where $\phi_{j}^{i,t}$ denotes the relevance score
 of client $j$ to $i$ in round $t$. In each round $t$, the models of clients with the top-$K$ relevance score are downloaded, forming a coalition $\mathcal{S}_{i,k}^{t}$ for client $i$; then, the SV $\varphi_{j}^{t}$ of each client $j\in \mathcal{S}_{i,k}^{t}$ is calculated and used to update the corresponding elements of $\phi^{i,t}$: $\phi_{j}^{i,t+1}=\eta \phi_{j}^{i,t} + (1-\eta)\varphi_{j}^{t}$. For client $i$, the weighted collaboration vector $\alpha_{i}$ is also set according to the SVs: $\alpha_{i,j}^{t}=\max\{\varphi_{j}^{t},\, 0\}/\lVert w_{i}^{t}-w_{j}^{t} \rVert$ if $j\in \mathcal{S}_{i,k}^{t}$ and $\alpha_{i,j}^{t}=0$ otherwise, where $\lVert \cdot \rVert$ denotes the Euclidean distance. The PFL scheme in \citep{Shapley-TMC-24} is called \textbf{pFedSV}.

\subsubsection{
Cosine Similarity}

Let $\mathbf{z_{i}}=(z_{i,1}, \dots, z_{i,d})$, where $i\in \{1,2\}$. The cosine similarity of two non-zero vectors is defined as $cos(\mathbf{z_{1}}, \mathbf{z_{2}}) = \frac{\mathbf{z_{1}}\cdot \mathbf{z_{2}}}{\lVert \mathbf{z_{1}} \rVert \lVert \mathbf{z_{2}} \rVert}$, where $\mathbf{z_{1}}\cdot \mathbf{z_{2}}$ is the inner product and $\lVert \cdot \rVert$ denotes the Euclidean norm. Intuitively, the closer the two vectors are, the smaller the angle between them, and thus the cosine value will be closer to one. Let us consider the personalized models $\mathbf{h_{i}}$ and $\mathbf{h_{j}}$ of clients $i$ and $j$. Their cosine similarity is $cos(\mathbf{h_{i}}, \mathbf{h_{j}})$. 

\vspace{0.1em}\noindent\textbf{Application.} Intuitively, when the data distribution of two clients $i$ and $j$ are more similar, their models are more similar, and their collaboration strength (i.e., $\alpha_{i,j}$) should be larger \citep{li2020federated,luo2021no}. 
The similarity between model parameters is used to guide the collaboration strength. In each training round $t$, \citet{Cosine-Similarity-ICML-23} proposes minimizing the following function in a privacy-preserving manner while adhering to the standard FL training process: $\sum\nolimits_{i=1}^{N}{{\beta_{i}\left( \mathcal{L}_{i}\left(\sum\nolimits_{j=1}^{N}{\alpha_{i,j}^{t}w_{j}^{t}}\right) - \frac{\lambda}{2}\sum\nolimits_{j=1}^{N}{\alpha_{i,j}^{t}cos(w_{i}^{t}, w_{j}^{t})} \right)}}$, where the decision variables are $\{w_{i}^{t}\}_{i=1}^{N}$ and $\alpha^{t}$, and $\lambda$ is a hyperparameter to balance the individual utilities and the collaboration necessity.


\subsubsection{
Hypernetworks}

Each client $i$ has a risk/loss function $\ell_{i}$: $\mathbb{R}^{N}\rightarrow \mathbb{R}_{+}$. Given a learned hypothesis $h\in$ $\mathcal{H}$, let the loss vector $\mathbf{\ell}(h)=[\ell_{1}, \dots, \ell_{N}]$ represent the utility loss of the $N$ clients under the hypothesis $h$. The hypothesis $h$ is considered a Pareto solution if there is no other hypothesis $h^{\prime}$ that dominates $h$, i.e., $\nexists h^{\prime}\in \mathcal{H}$, s.\,t.\, $\forall i: \ell_{i}(h^{\prime})\leqslant \ell_{i}(h)$ and $\exists j: \ell_{j}(h^{\prime}) < \ell_{j}(h)$. Let $r=(r_{1}, \dots,r_{N})$ $\in \mathbb{R}^{N}$ denote a preference vector where $\sum_{k=1}^{N}{r_{k}}=1$ and $r_{k}\geq 0, \forall k\in \{1, \dots, N\}$. The hypernetwork $HN$ takes $r$ as input and outputs a Pareto solution $h$, i.e., $h\gets HN(\phi, r)$, where $\phi$ denotes the parameters of the hypernetwork \citep{navon2020learning}.  

\vspace{0.05em}\noindent\textbf{Application.} For each client $i$, linear scalarization can be used. \citet{cui2022collaboration} determine an optimal preference vector $r_{i}^{\ast}$ to generate the hypothesis $h_{i}^{\ast}$ that minimizes the loss with the data $\hat{\mathcal{D}}_{i}$. This is expressed as $h_{i}^{\ast}=HN(\phi, r_{i}^{\ast})$, where $r_{i}^{\ast} = \argmin\nolimits_{r}{\hat{\mathcal{L}}_{i}(HN(\phi, r))}$. 
For each client $i$, the value of $r_{i}^{\ast}$ is assigned to $\alpha_{i}$. In \citep{cui2022collaboration}, collaboration equilibrium (CE) is sought with additional considerations. We refer to the PFL scheme here as \textbf{CE} due to its origin in \citep{cui2022collaboration}.

\begin{table}[b!]
\centering
\caption{Comparison of various schemes under different conditions}
\label{tab:comparison}
\resizebox{\textwidth}{!}{%
\begin{tabular}{ccccccccc}
\toprule
\textbf{Category} & \textbf{Dataset} & \textbf{Partitioning} & \textbf{pFedGraph} & \textbf{pFedSV} & \textbf{pFedJS} & \textbf{FedCollab} & \textbf{RACE} & \textbf{CE}\\ \midrule
\multirow{22}{*}{\parbox{2.5cm}{\centering \textbf{Label}\\\textbf{distribution}\\\textbf{skew}}} 
 & MNIST & $p_k \sim Dir(0.5)$ & \textbf{98.90\%$\pm$0.10\%} & 91.18\%$\pm$3.84\% & 98.54\%$\pm$0.40\% & 79.66\%$\pm$3.59\% & \textbf{98.90\%$\pm$0.27\%} & 98.82\%$\pm$0.15\% \\
 &  & \#C=1 & \textbf{10.65\%$\pm$0.70\%}  & 10.24\%$\pm$0.08\% & 10.09\%$\pm$0.00\% & 10.00\%$\pm$0.00\% & 10.09\%$\pm$0.00\% &10.30\%$\pm$0.02\%\\
 &  & \#C=2 & \textbf{95.60\%$\pm$0.43\%} & 35.52\%$\pm$6.90\% & 10.17\%$\pm$0.15\% &  19.93\%$\pm$0.47\%& 10.17\%$\pm$0.15\% &84.65\%$\pm$7.36\% \\
 &  & \#C=3 & \textbf{97.34\%$\pm$0.81\%} &52.95\%$\pm$16.93\%  & 9.84\%$\pm$0.92\% & 27.96\%$\pm$1.67\% & 12.73\%$\pm$6.26\% &93.52\%$\pm$2.34\% \\ \cmidrule(lr){2-9}
 & FMNIST & $p_k \sim Dir(0.5)$ & \textbf{87.54\%$\pm$0.44\%} & 80.74\%$\pm$2.25\% & 85.34\%$\pm$1.53\% & 69.10\%$\pm$2.01\% & 85.68\%$\pm$1.51\% & 86.93\%$\pm$0.83\% \\
 &  & \#C=1 &\textbf{10.67\%$\pm$1.33\%}  & 10.05\%$\pm$0.09\% & 10.00\%$\pm$0.00\% & 10.00\%$\pm$0.00\% & 10.00\%$\pm$0.00\% &10.00\%$\pm$0.00\% \\
 &  & \#C=2 & \textbf{71.37\%$\pm$8.10\%} & 26.94\%$\pm$1.29\% & 10.00\%$\pm$0.00\% & 18.81\%$\pm$1.19\%  & 10.00\%$\pm$0.00\% &58.44\%$\pm$6.91\% \\
 &  & \#C=3 & \textbf{80.96\%$\pm$2.21\%} & 34.44\%$\pm$5.55\% & 10.00\%$\pm$0.00\% &26.56\%$\pm$1.03\%  & 25.45\%$\pm$30.90\% & 74.11\%$\pm$1.79\%\\ \cmidrule(lr){2-9}
 & CIFAR-10 & $p_k \sim Dir(0.5)$ & \textbf{64.69\%$\pm$0.92\%} & 44.87\%$\pm$1.48\% & 45.80\%$\pm$1.68\% & 36.10\%$\pm$0.51\% & 49.32\%$\pm$3.52\% & 62.65\%$\pm$3.00\%\\
 &  & \#C=1 & 10.00\%$\pm$0.00\% & \textbf{10.08\%$\pm$0.16\%} & 10.00\%$\pm$0.00\% &  10.00\%$\pm$0.00\%&  10.00\%$\pm$0.00\%&10.00\%$\pm$0.00\%\\
 &  & \#C=2 & 39.77\%$\pm$7.50\% & 17.53\%$\pm$6.53\% & 10.00\%$\pm$0.00\% & 17.20\%$\pm$0.98\% & 10.00\%$\pm$0.00\% &\textbf{42.87\%$\pm$6.35\%}\\
 &  & \#C=3 & \textbf{56.00\%$\pm$0.03\%} & 26.33\%$\pm$4.99\% & 10.00\%$\pm$0.00\% &23.78\%$\pm$0.19\%  & 10.00\%$\pm$0.00\% &49.27\%$\pm$1.03\%\\ \cmidrule(lr){2-9}
 & SVHN & $p_k \sim Dir(0.5)$ & 86.59\%$\pm$0.39\% & 63.23\%$\pm$6.19\% & 71.11\%$\pm$7.10\% & 54.93\%$\pm$3.96\% & 74.85\%$\pm$5.48\% & \textbf{87.64\%$\pm$2.33\%}\\
 &  & \#C=1 & \textbf{18.37\%$\pm$2.43\%} & 11.29\%$\pm$0.00\% & 6.13\%$\pm$0.00\% & 10.00\%$\pm$0.00\% & 6.13\%$\pm$0.00\% & 9.03\%$\pm$1.55\%\\
 &  & \#C=2 & \textbf{69.35\%$\pm$6.51\%} & 27.63\%$\pm$5.98\% & 15.53\%$\pm$4.46\% & 18.59\%$\pm$3.01\% & 15.53\%$\pm$4.46\% &52.22\%$\pm$12.74\%\\
 &  & \#C=3 & 70.29\%$\pm$2.32\% & 51.56\%$\pm$4.39\% & 10.95\%$\pm$4.99\% & 23.11\%$\pm$1.88\% & 10.95\%$\pm$4.99\% &\textbf{70.47\%$\pm$1.77\%}\\ \cmidrule(lr){2-9}
 & adult & $p_k \sim Dir(0.5)$ & \textbf{80.83\%$\pm$0.83\%} & 78.97\%$\pm$3.57\% & 76.40\%$\pm$0.05\% & 72.89\%$\pm$5.35\%& 77.57\%$\pm$2.29\% & 74.86\%$\pm$5.15\%\\
 &  & \#C=1 & \textbf{67.58\%$\pm$17.58\%} & 39.45\%$\pm$0.00\% & 23.62\%$\pm$0.00\% & 50.00\%$\pm$0.00\% & 23.62\%$\pm$0.00\% &50.00\%$\pm$0.00\%\\ \cmidrule(lr){2-9}
 & rcv1 & $p_k \sim Dir(0.5)$ & 49.66\%$\pm$0.82\% & 61.02\%$\pm$10.36\% & 49.35\%$\pm$1.81\% & 62.00\%$\pm$10.27\% & 49.59\%$\pm$2.06\% &\textbf{79.55\%$\pm$5.58\%}\\
 &  & \#C=1 & \textbf{51.16\%$\pm$0.00\%} & 50.28\%$\pm$0.05\% & \textbf{51.16\%$\pm$0.00\%} & 50.00\%$\pm$0.00\% & 51.16\%$\pm$0.00\% &50.00\%$\pm$0.00\% \\ \cmidrule(lr){2-9}
 & covtype & $p_k \sim Dir(0.5)$ & 50.72\%$\pm$0.48\% & 50.96\%$\pm$0.24\% & 50.00\%$\pm$1.20\% &50.24\%$\pm$0.00\% & \textbf{51.16\%$\pm$0.00\%} & 50.58\%$\pm$0.58\%\\
 &  & \#C=1 & \textbf{50.00\%$\pm$0.00\%} & 49.28\%$\pm$0.48\% & 48.80\%$\pm$0.00\% & \textbf{50.00\%$\pm$0.00\%} & 48.80\%$\pm$0.00\% &\textbf{50.00\%$\pm$0.01\%}\\ \midrule
\multicolumn{3}{l}{\textbf{Number of times that performs the best}} & 16 & 1 & 1 & 1 & 1 & 5\\ \midrule
\multirow{5}{*}{\parbox{2.5cm}{\centering \textbf{Feature}\\\textbf{distribution}\\\textbf{skew}}} 

 & MNIST & \multirow{4}{*}{\textbf{$\mathbf{\hat{x}} \sim Gau(0.1)$}} &98.78\%$\pm$0.22\%  & 98.87\%$\pm$0.25\% & 99.02\%$\pm$0.10\% & 98.61\%$\pm$0.06\% & \textbf{99.11\%$\pm$0.01\%} &98.98\%$\pm$0.06\%\\
 & FMNIST && 88.50\%$\pm$0.01\%  & 84.13\%$\pm$0.79\%  & \textbf{88.61\%$\pm$0.00\%} & 87.37\%$\pm$0.40\% & 87.77\%$\pm$0.01\% &88.43\%$\pm$0.41\%\\
 & CIFAR-10 && 69.18\%$\pm$0.80\% & 62.24\%$\pm$0.06\% & 70.09\%$\pm$0.54\% & 65.82\%$\pm$3.41\% & \textbf{70.96\%$\pm$0.17\%} & 67.44\%$\pm$1.93\% \\
 & SVHN && 87.18\%$\pm$2.71\% & 85.65\%$\pm$0.09\% & 87.34\%$\pm$0.57\% & 86.29\%$\pm$0.00\% & 86.57\%$\pm$0.28\%& \textbf{87.61\%$\pm$0.03\%} \\ \cmidrule(lr){2-9}
\multirow{1}{*}{} & FEMNIST & real-world & 99.27\%$\pm$0.04\% & 97.80\%$\pm$1.15\% & 99.23\%$\pm$0.00\% &98.93\%$\pm$0.20\%  & \textbf{99.28\%$\pm$0.01\%} &99.26\%$\pm$0.01\%\\ \midrule
\multicolumn{3}{l}{\textbf{Number of times that performs the best}} & 0 & 0 & 1 & 0 & 3 & 1\\ \midrule
\multirow{6}{*}{\parbox{2.5cm}{\centering \textbf{Quantity}\\\textbf{skew}}}

 & MNIST & \multirow{6}{*}{\textbf{$q \sim Dir(0.5)$}} & 99.04\%$\pm$0.06\% & 98.29\%$\pm$0.67\% & 99.12\%$\pm$0.12\% &96.61\%$\pm$3.43\% & \textbf{99.14\%$\pm$0.08\%} & 98.99\%$\pm$0.04\%\\
 & FMNIST && \textbf{88.85\%$\pm$0.29\%} & 84.78\%$\pm$3.92\% &  88.31\%$\pm$0.30\%& 87.10\%$\pm$1.12\% &87.90\%$\pm$0.28\% &88.48\%$\pm$0.20\%   \\
 & CIFAR-10 && \textbf{72.19\%$\pm$0.03\%} & 61.68\%$\pm$0.98\% &71.70\%$\pm$0.40\%  & 62.51\%$\pm$2.91\% & 71.86\%$\pm$0.35\% & 70.12\%$\pm$0.70\% \\
 & SVHN &  & 88.16\%$\pm$0.17\% & 76.63\%$\pm$1.52\% & 85.05\%$\pm$0.72\% & 85.14\%$\pm$0.44\% & 84.54\%$\pm$0.26\% &\textbf{88.52\%$\pm$0.39\%}\\
 & adult &  & \textbf{82.34\%$\pm$0.04\%} & 82.24\%$\pm$1.26\% & 82.23\%$\pm$0.08\% & 82.15\%$\pm$0.18\% & 82.25\%$\pm$0.29\% &82.07\%$\pm$0.10\%\\
 & rcv1 &  & 51.16\%$\pm$0.00\% & 51.40\%$\pm$0.25\% & \textbf{96.31\%$\pm$0.17\%} & 95.44\%$\pm$0.31\%& 95.73\%$\pm$0.57\% & 96.25\%$\pm$0.22\%\\
 & covtype &  & \textbf{51.20\%$\pm$0.00\%} &  \textbf{51.20\%$\pm$0.00\%}& \textbf{51.20\%$\pm$0.10\%}& \textbf{51.20\%$\pm$0.00\%}& \textbf{51.20\%$\pm$0.00\%} & \textbf{51.20\%$\pm$0.05\%}\\ \midrule
\multicolumn{3}{l}{\textbf{Number of times that performs the best}} & 4 & 1 & 2 & 1 & 2 &1\\ \midrule
\multirow{7}{*}{\parbox{2.5cm}{\centering \textbf{Homogeneous}\\\textbf{partition}}}
 & MNIST & \multirow{7}{*}{\textbf{IID}} & 98.96\%$\pm$0.04\% & 98.63\%$\pm$0.07\% & 99.06\%$\pm$0.12\% & 98.60\%$\pm$0.00\% & \textbf{99.13\%$\pm$0.00\%} &98.97\%$\pm$0.04\%\\
 & FMNIST &  & \textbf{88.91\%$\pm$0.03\%} & 82.36\%$\pm$0.71\% & 87.77\%$\pm$0.07\% & 87.30\%$\pm$0.41\% & 88.16\%$\pm$0.04\% &88.36\%$\pm$0.42\%\\
 & CIFAR-10 &  & 68.10\%$\pm$0.44\% & 62.18\%$\pm$0.57\% & 69.59\%$\pm$0.57\% &  63.28\%$\pm$0.88\%& \textbf{70.39\%$\pm$0.17\%} &66.54\%$\pm$1.81\%\\
 & FEMNIST &  & \textbf{99.33\%$\pm$0.01\%} & 97.64\%$\pm$1.36\% & 99.24\%$\pm$0.03\% & 99.12\%$\pm$0.06\%& 99.28\%$\pm$0.02\% &99.25\%$\pm$0.00\%\\
 & SVHN &  & \textbf{88.41\%$\pm$0.14\%} & 85.71\%$\pm$0.04\% & 86.07\%$\pm$0.35\% &  87.25\%$\pm$1.24\%& 86.82\%$\pm$0.23\% &87.84\%$\pm$0.35\%\\
 & adult &  & 82.22\%$\pm$0.29\% & 81.37\%$\pm$0.55\% & 81.91\%$\pm$1.09\% & 82.53\%$\pm$0.47\%& 81.76\%$\pm$0.39\% &\textbf{83.87\%$\pm$2.04\%} \\
 & rcv1 &  &  51.40\%$\pm$0.25\%& 95.28\%$\pm$0.21\% & \textbf{96.15\%$\pm$0.38\%} & 51.40\%$\pm$0.25\%& 95.88\%$\pm$0.32\% & 81.11\%$\pm$29.95\%\\
 & covtype &  & \textbf{51.23\%$\pm$0.06\%} & 51.20\%$\pm$0.01\% & \textbf{51.23\%$\pm$0.06\%} & 51.25\%$\pm$0.05\% & 49.95\%$\pm$1.25\% &\textbf{51.23\%$\pm$0.06\%}\\ \midrule
\multicolumn{3}{l}{\textbf{Number of times that performs the best}} & 4 & 0 & 2 & 0 & 2 &2\\ \bottomrule
\end{tabular}%
}
\end{table}

\section{Experimental Studies}
\label{others}

To investigate the effectiveness of existing FL schemes in studying the statistical heterogeneity of data, we conducted extensive experiments on eight public datasets. These included five image datasets (i.e., MNIST \citep{lecun1998gradient}, CIFAR-10 \citep{krizhevsky2009learning}, FMNIST \citep{xiao2017fashion}, SVHN \citep{netzer2011reading}, FEMNIST \citep{caldas2018leaf}), and three tabular datasets (i.e., adult \citep{kohavi1996scaling}, rcv1 \citep{lewis2004rcv1}, and covtype \citep{blackard1999comparative}). For specific experimental procedures, we followed the protocol outlined in \citep{NonIID-Benchmarking-ICDE-22}. By default, the framework involves ten clients, each performing training over 10 epochs per communication round with a batch size of 64, across a total of 50 communication rounds. 
We use top-1 accuracy to indicate the differences in test case accuracy of various schemes under different non-IID conditions. Unless otherwise specified, the value in the tables represent the average test case results across all clients using their personalized models. For more details on the experimental setup and datasets, please refer to Appendix \ref{append-exp-setup}. The main insights of this paper are highlighted in {\em italic text}.

 
\subsection{Analysis of the Accuracy obtained by Collaboration}

The accuracy of the six methods under the five standard non-IID settings and the IID data setting is shown in Table \ref{tab:comparison}. Non-IID data can degrade the FL performance to some extent. The results under the IID setting is used as a baseline to evaluate the performance of the six methods under non-IID settings. 
\begin{figure}[t]
    \centering
    \includegraphics[width=0.88\textwidth]{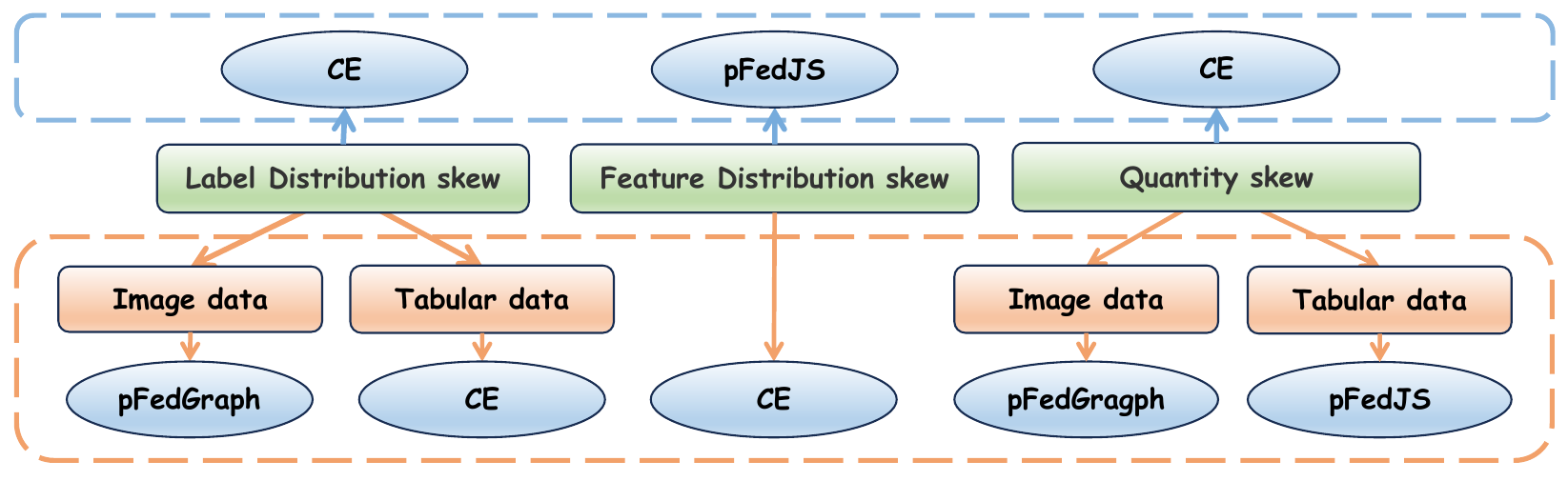}
    \caption{The decision tree for identifying the optimal FL scheme under different non-IID setting.}
    \label{fig:decision-trees}
\end{figure}

\vspace{0.1em}\noindent\textbf{Performance Comparison.}
Figure \ref{fig:decision-trees} summarizes the results in Table \ref{tab:comparison}, with the blue ovals at the bottom illustrating which schemes are advantageous under each setting. There is no universally best scheme; the choice depends on the specific setting.
In the label distribution skew setting, {\em pFedGraph} significantly outperforms other schemes on image data, while {\em CE} performs best on average for tabular data. For feature distribution skew, {\em RACE} usually achieves the highest accuracy. In the quantity skew setting, {\em pFedGraph} continues to excel on image data, while {\em pFedJS} achieves the highest accuracy on tabular data. Our observation is as follows: {\em as illustrated by the bottom blue ovals of Figure \ref{fig:decision-trees}, each non-IID setting has a corresponding technique that maximizes the collaboration advantages and personalized model accuracy.}

The pairwise collaboration advantages of clients can be measured by the matrix $\alpha^{t}$ or $\alpha$, which defines a benefit graph of clients. In the line of research \citep{cui2022collaboration,tan2024fedcompetitors,Wu22a,Chen2024FedEgoists}, some desired properties (i.e., additional constraints) need to be maintained in the collaboration relationships of clients to address the issues with fairness, collaboration and competition \citep{lyu2020collaborative,Lyu20a,Huang24a}; then, the final collaboration relationships of clients will form a subgraph of the benefit graph. Currently, only the hypernetwork technique is used to estimate $\alpha^{t}$, without considering whether other techniques might be more advantageous. 
As identified by this paper, the matrix $\alpha^{t}$ can in fact be estimated by the {\em pFedJS}, {\em FedCollab}, {\em pFedSV}, {\em pFedGraph}, and {\em hypernetwork} techniques, where hypernetwork is used in the CE scheme. In the case of \citep{cui2022collaboration}, the matrix $\alpha^{t}$ needs to be computed and the collaboration relationships of clients are determined before the FL training process starts; once these relationships are established, they remain fixed during the entire training process; thus, a fixed matrix $\alpha$ is required, and only the {\em pFedJS}, {\em FedCollab}, and {\em Hypernetwork} techniques can be applied. In the case of \citep{tan2024fedcompetitors}, both the matrix $\alpha^{t}$ and the collaboration relationships of clients can be updated in each round of FL training, as long as it does not create conflicts of interest among clients; thus, all five techniques mentioned above can be used to compute the matrix $\alpha^{t}$. Regarding the choice of proper techniques for generating the benefit graph, the better technique is the one achieving higher overall accuracy. From Table \ref{tab:comparison}, we make the following observations: (\rmnum{1}) {\em In the case where the benefit graph (i.e., the matrix $\alpha^{t}$) can be updated in each round of FL training, it is desirable that under each non-IID setting, the matrix $\alpha^{t}$ is generated by the corresponding technique illustrated by the blue ovals at the bottom of Figure \ref{fig:decision-trees}. } (\rmnum{2}) {\em In the case where the benefit graph (i.e., the matrix $\alpha$) needs to be fixed and computed before the FL training process starts, it is desirable that under each non-IID setting, the matrix $\alpha$ is generated by the corresponding technique illustrated by the blue ovals at the top of Figure \ref{fig:decision-trees}.}

\vspace{0.1em}\noindent\textbf{Effect of different non-IID settings.} It can be observed from Table \ref{tab:comparison} that: 1) for the feature distribution and quantity skew settings, existing schemes achieve accuracy close to the IID setting; 2) among all non-IID settings, the label distribution skew is the most challenging. Notably, in the quantity-based label imbalance setting, as the number of classes $C$ decreases from three to one, the degree of data heterogeneity increases, leading to a decrease in accuracy. In the future, {\em there is still room to improve existing schemes for the label distribution skew setting}.

\vspace{0.1em}\noindent\textbf{Effect of different tasks.} 
CIFAR-10 is the most complex dataset, and its accuracy varies significantly across different schemes and non-IID settings. For other, simpler image datasets, the accuracy differences are relatively small across all cases. Tabular data also presents significant challenges, with accuracy consistently low across all six schemes (e.g.,  $\leqslant 52\%$ for covtype in all cases). These observations suggest that {\em more challenging tasks like CIFAR-10 and tabular datasets should be included in future benchmarks to better study the complementarity among clients}.

Finally, as generalized in Eq. (\ref{equa-nju}), the difference between the weighted empirical risk and the optimal expected risk can be well bounded when two types of divergences are applied, respectively. {\em An interesting direction for future research is to explore whether such a bound exists for other divergence measures, such as the earth mover distance \citep{andoni2008earth} and the Fr\'echet distance \citep{heusel2017gans}, which are mentioned in \citep{DSketch-NIPS-23}.}

\begin{table}[t]
\centering
\caption{The computation time (seconds) and communication size (MBs) of different schemes.}
\label{tab:efficiency}
\footnotesize
\resizebox{0.9\textwidth}{!}{%
\begin{tabular}{lcccccccc}
\toprule
 & \multicolumn{4}{c}{\textbf{Computation Time}} & \multicolumn{4}{c}{\textbf{Communication Size}} \\ \cmidrule(lr){2-5} \cmidrule(lr){6-9}
 & \textbf{MNIST} & \textbf{CIFAR-10} & \textbf{adult} & \textbf{rcv1} & \textbf{MNIST} & \textbf{CIFAR-10} & \textbf{adult} & \textbf{rcv1} \\ \midrule
\textbf{pFedJS} & 23.69s & 124.1s & 1.591s & 30.87s & {170.1MB} & {240.1MB} & {20.1MB} & {5770.1MB} \\ 
\textbf{pFedGraph} & 9.410s & 6.340s & 4.870s & 27.02s & 170MB & 240MB & 20MB & 5770MB \\
\textbf{pFedSV} & 1258s & 5444s & 149.1s & 244.5s & 850MB & 1200MB & 100MB & 28850MB \\
\textbf{FedCollab} & 305.9s & 1312s & 48.88s & 63.69s & 215.9MB & 304.8MB & 25.4MB & 7327.9MB \\
\textbf{RACE} & 4.169s & 27.26s & 2.003s & 3.017s & 105.9MB & 145.2MB & 13.3MB & 3462.6MB \\
\textbf{CE} & 391.8s & 1740s & 142.4s & 166.5s & 33330.1MB & 47020.1MB & 3490.1MB & 925820.1MB \\
\bottomrule
\end{tabular}
}%
\end{table}

\subsection{Efficiency}

Schemes except for \citep{DSketch-NIPS-23} compute a matrix \(\alpha\) only once before the FL training process starts \citep{cui2022collaboration,Ding-NIPS-22,FedCollab-ICML-23} or a matrix \(\alpha^{t}\) in each round of FL training \citep{Cosine-Similarity-ICML-23,Shapley-TMC-24} to quantify the collaboration advantages of clients. \citep{DSketch-NIPS-23} performs a one-time computation of the distance/divergence between the client and global data distribution sketches before the FL training process starts. 
We evaluated (\rmnum{1}) the time required to compute the matrix (\(\alpha^{t}\) or $\alpha$) or the divergence in each scheme under the noise-based feature imbalance setting, and (\rmnum{2}) the communication overhead with the server (i.e., the amount of data communicated with the server over 50 communication rounds). Table \ref{tab:efficiency} shows the experimental results. 

The computational overhead of pFedSV is the highest across all datasets, followed closely by CE and FedCollab. The high computational overhead of these schemes is due to their use of neural networks to compute \(\alpha\). In pFedJS, the JS divergence is computed in the joint feature and label space, whose dimension can be high; we note that the implementation in \citep{Ding-NIPS-22} considers only the label distribution skew case, where only the label space is concerned. In pFedGraph, the optimization of personalized models is performed locally. RACE and pFedJS execute simple binary operations, resulting in significantly faster computation. 
For the CE scheme, all clients train a Hypernetwork, resulting in a large amount of data exchanged with the server during the preparation of \(\alpha\). The communication overhead for pFedGraph, pFedJS, FedCollab and RACE is similar, as they typically require uploading the local model and downloading the personalized model during training. However, pFedSV has a relatively higher communication overhead because it needs to compute the Shapley Value locally using other clients' models.

Recently, there has been growing interest in measuring collaboration advantages to establish collaboration relationships among clients in the Internet of Things (IoT) field \citep{Coalitional-FL-TMC-24,Guo-Song-23a,zhu2024shufflefl,lu2024leap,Coalitional-FL-TSC-23}. In these works, the JS divergence is commonly used. As pointed out in \citep{DSketch-NIPS-23}, divergence estimation can be difficult in high-dimensional spaces and require expensive computation \citep{huang2015heads,andoni2008earth}. 
In IoTsscenarios, clients are typically resource-constrained devices with limited hardware resources (energy, bandwidth, computation). With these considerations, \citet{DSketch-NIPS-23} propose a one-time distribution sketch technique to estimate data distribution divergence using lightweight computation. However, in their scheme, \citep{DSketch-NIPS-23} don't need to generate the matrix $\alpha$ or $\alpha^{t}$ to quantify the collaboration advantages of clients. As shown in Table \ref{tab:efficiency}, using distribution sketches to measure divergence is much faster than using JS divergence. On the other hand, FL is often studied from a game theory perspective \citep{tu2022incentive,gupta2023federated,khan2021federated}, where each client in the collaborative FL network aims to maximize its utility by collaborating with others \citep{donahue2021model}. Thus, the collaboration advantages (i.e., $\alpha$ or $\alpha^{t}$) of clients need to be well understood and quantified in a lightweight way for the IoTsapplications. For example, \citet{lu2024leap} consider applying FL to IoTs and assume that the collaboration advantage (i.e., the utility) is linearly proportional to the JS divergence, which may be a simplistic assumption. Thus, {\em to make the existing technical development in literature more practical, it may be promising to study how to use the lightweight one-time distribution sketch technique to generate the matrix $\alpha$ or $\alpha^{t}$}, which lays a practical foundation for the FL collaboration establishment in the IoTs field.

\begin{figure}[h!]
\centering
\includegraphics[width=\linewidth]{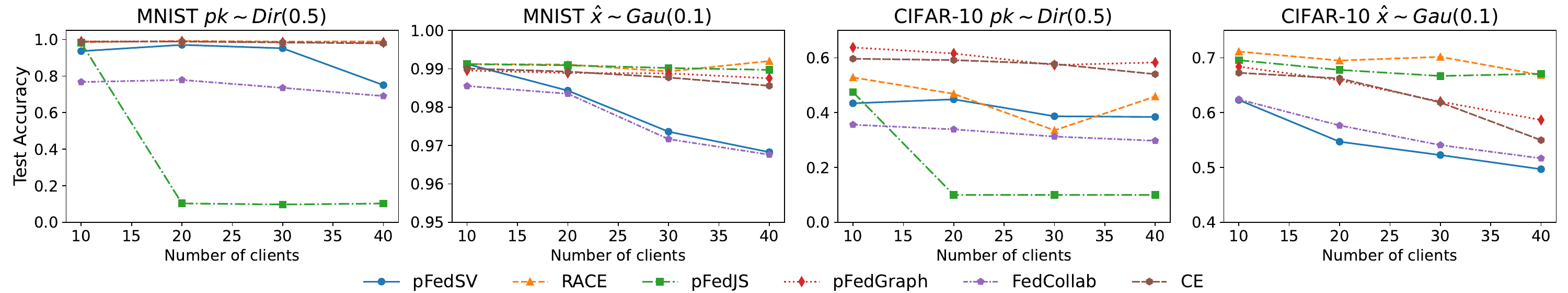}
\caption{The change of test accuracy with the number of clients: the first (resp. last) two plots are for CIFAR-10 (resp. MNIST).}
\label{fig:num-parties}
\end{figure}

\subsection{Other Experiments}


\vspace{0.1em}\noindent\textbf{Scalability.} From Figure \ref{fig:num-parties}, it can be observed that the accuracy of schemes may vary, and in many cases, their accuracy decreases as the number of clients increases. Among the six schemes, no scheme dominates the other schemes across all cases. The sensitivity of these schemes' performance to the number of clients implies that scalability should be an important aspect to consider when evaluating a FL scheme in the related research. 
\begin{table}[h]
\centering
\caption{The performance of different schemes with mixed imbalance cases on CIFAR-10.}
\label{tab:performance}
\resizebox{0.76\textwidth}{!}{
\begin{tabular}{lcccccc}
\toprule
 & \textbf{pFedGraph} & \textbf{pFedSV} & \textbf{pFedJS} & \textbf{FedCollab} & \textbf{RACE} & \textbf{CE}\\ 
\midrule
Feature and quantity skew & 31.38\% & 21.67\% & 27.78\% & 19.21\% & 10.00\%& 28.26\%\\
\midrule
Label and feature skew & 10.51\% & 20.73\% & 10.00\% & 11.89\% & 10.00\%& 10.00\%\\
\bottomrule
\end{tabular}
}
\end{table}

\vspace{0.1em}\noindent\textbf{Mixed Types of Data Skew.} In reality, different types of skew may coexist. Like \citep{NonIID-Benchmarking-ICDE-22}, we consider two settings: mixed label and feature skew, and mixed feature and quantity skew; please refer to Appendix \ref{append-mixed-skew} for details. The experimental results are presented in Table \ref{tab:performance}. 
When facing mixed types of skew, all these schemes encounter significant challenges. For example, in the mixed feature and quantity skew setting, the performance of all schemes on CIFAR-10 is poor, with accuracy not exceeding 32\%,  much lower than the case when there is only one type of skew, as illustrated in Table \ref{tab:comparison}. 
{\em It is crucial to find effective solutions for the mixed types of skew case since this case is also common in real-world scenarios.}

The effects of batch size, communication rounds, and local epochs on these PFL schemes have also been evaluated through experiments. Please refer to Appendix \ref{append:config-FL-training-process} for detailed results and analysis.

\section{Concluding Remarks}

In this paper, we summarize the six major techniques (JS divergence, $\mathcal{C}$-divergence, distribution sketch, Shapley Value, Hypernetworks, cosine similarity) to quantify data heterogeneity in FL settings into a first-of-its-kind unified framework to understand their effects in-depth. Extensive experiments over eight popular datasets have been conducted to compare these schemes under five standard non-IID FL settings, providing much-needed insight into which schemes are advantageous under which settings. The unified framework and the experimental results identify the scenarios, under which the current schemes perform relatively poorly, and future research problems for PFL. It is useful for identifying the right techniques for quantifying the collaboration advantages among clients, guiding the related research on collaboration, fairness, and competition in FL settings. The findings suggest that lightweight FL schemes based on techniques such as the distribution sketch are worth studying in the future.


\appendix

\section{Experimental Setup Details}
\label{append-exp-setup}

The statistics of the datasets is summarized in Table \ref{tab:datasummary}. For image datasets, we employed a CNN with two 5x5 convolutional layers followed by 2x2 max pooling, and two fully connected layers with ReLU activation \citep{HE2024104129}. For tabular datasets, we used an MLP with three hidden layers for training. We used the SGD optimizer with a learning rate of 0.1 for rcv1 and 0.01 for the other datasets, along with a momentum of 0.9.

\begin{table}[htbp]
\centering
\caption{The summary of the Datasets in the experiment}
\label{tab:datasummary}
\resizebox{\textwidth}{!}{%
\begin{tabular}{lcccccc}
\toprule
&Datasets & \#training cases & \#validation cases & \#test cases & \#features & \#classes \\
\midrule
\multirow{5}{*}{\parbox{2.5cm}{\centering \textbf{Image}\\\textbf{Datasets}}} 
&MNIST & 48,000 & 12,000 & 10,000& 784 & 10 \\
&FMNIST & 48,000 & 12,000 & 10,000& 784 & 10 \\
&CIFAR-10 & 40,000 & 10,000 & 10,000& 1,024 & 10 \\
&SVHN &58606& 14651 & 26,032 & 1,024 & 10 \\
&FEMNIST & 273,498&68,375& 40,832 & 784 & 10 \\ \midrule
\multirow{3}{*}{\parbox{2.5cm}{\centering \textbf{Tabular}\\\textbf{Datasets}}} 
&adult & 26,049&6,512& 16,281 & 123 & 2 \\
&rcv1 &12,146&3,036& 5,060 & 47,236 & 2 \\
&covtype &348,607&87,152& 145,253 & 54 & 2 \\

\bottomrule
\end{tabular}
}
\end{table}

\section{Mixed Types of Skew}
\label{append-mixed-skew}

In the mixed label and feature skew setting, the distribution-based label imbalanced partitioning strategy and the noise-based feature imbalance strategy are applied sequentially. Specifically, the whole dataset is first divided into each client using the distribution-based label imbalanced partitioning strategy. Then, noise is added to the data of each client according to the noise-based feature imbalance strategy. As a result, both label distribution skew and feature distribution skew exist among the local data of different clients.

In the mixed feature and quantity skew setting, the quantity imbalanced partitioning strategy and the noise-based feature imbalance strategy are applied sequentially. Specifically, the whole dataset is allocated into each client by the quantity imbalanced partitioning strategy. Then, noise is added to the data of each client through the noise-based feature imbalance strategy. Consequently, both feature distribution skew and quantity skew exist among the local data of different clients.

\section{FL Training Process}
\label{append:config-FL-training-process}


\subsection{Communication Efficiency}

\begin{figure}[htp]
    \centering
    \includegraphics[width=0.9\linewidth]{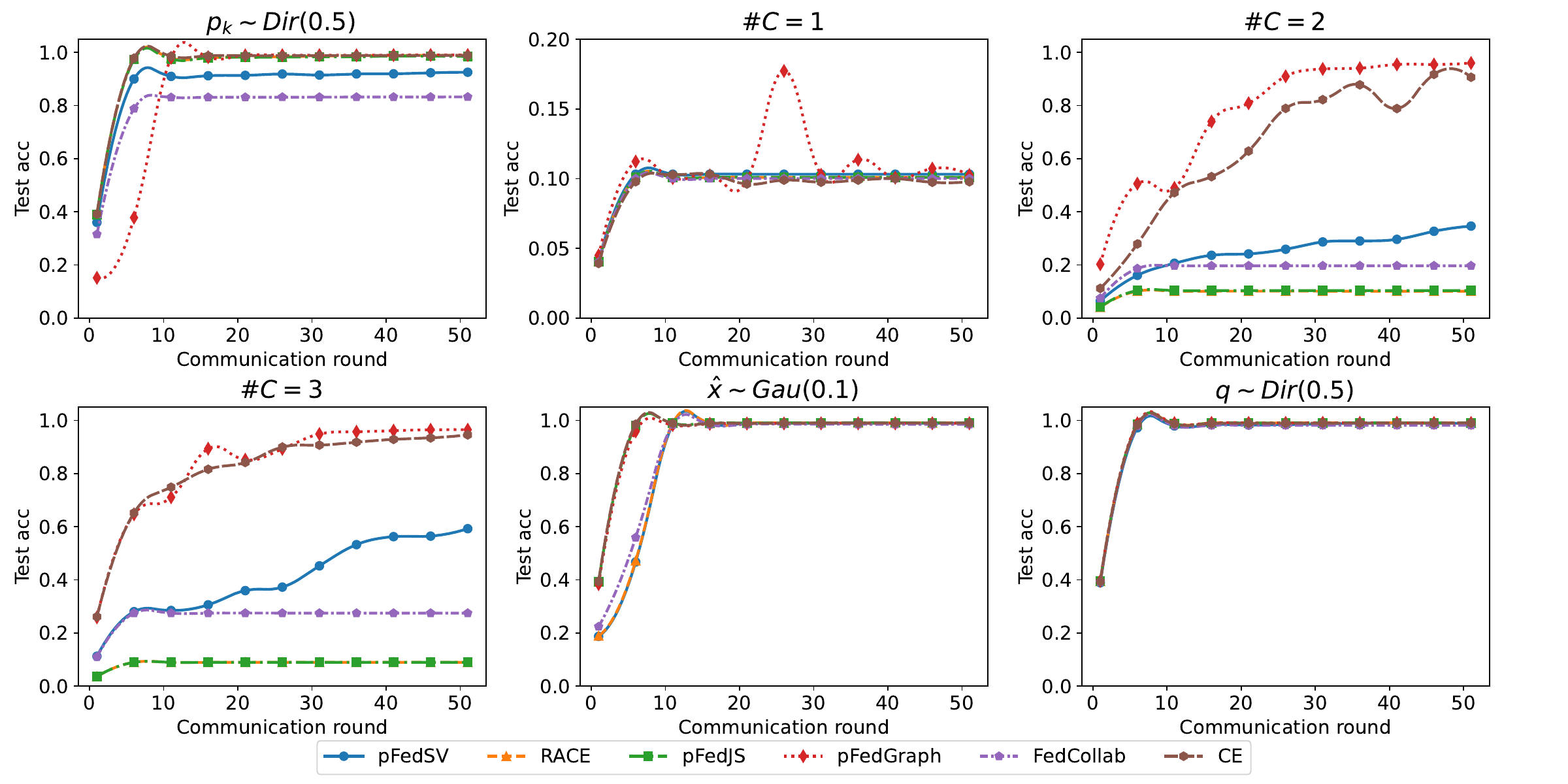}
    \caption{The training curves of different approaches on MNIST.}
    \label{fig:communication-efficiency-1}
\end{figure}

\begin{figure}[htp]
    \centering
    \includegraphics[width=0.9\linewidth]{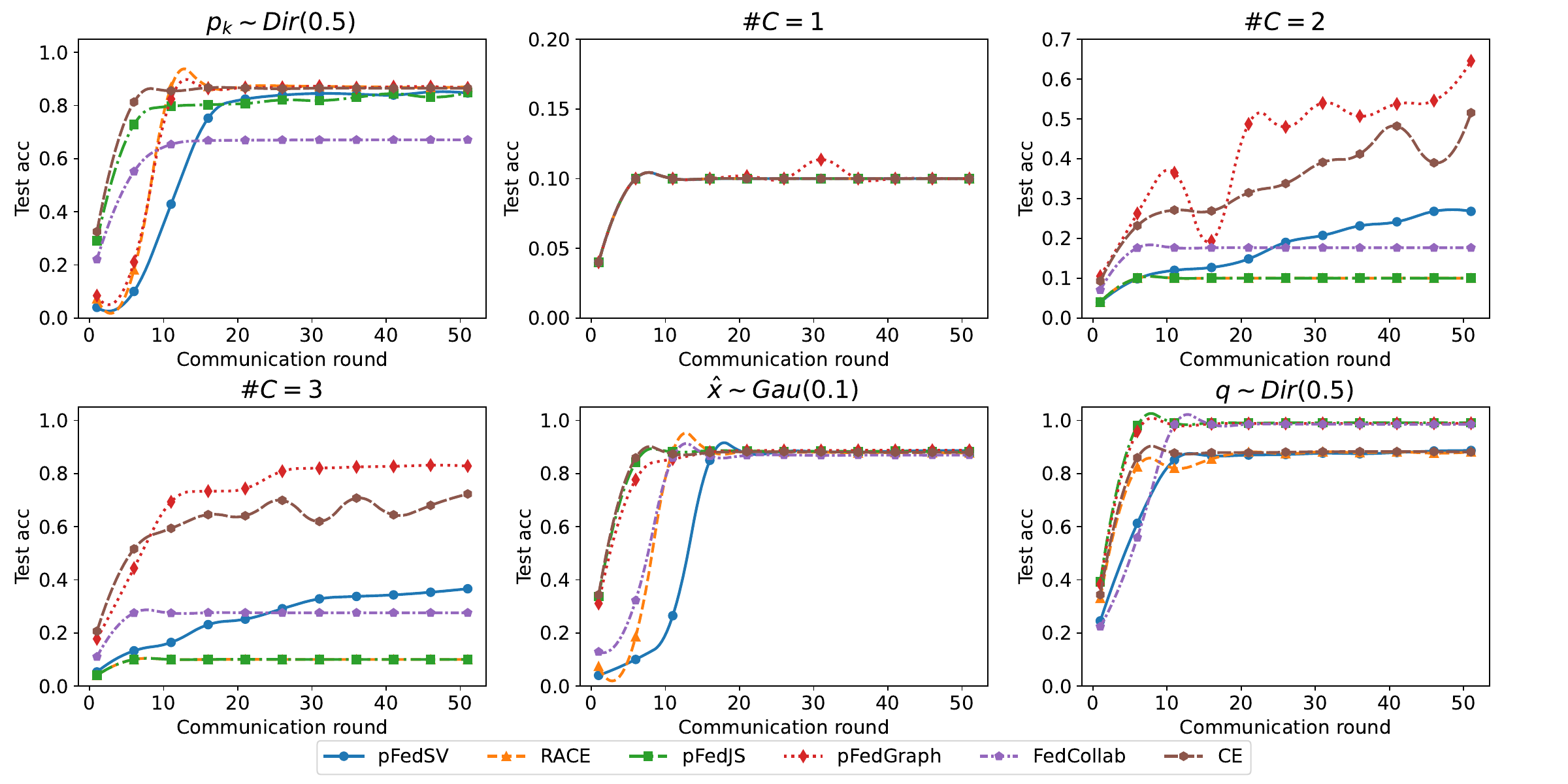}
    \caption{The training curves of different approaches on FMNIST.}
    \label{fig:communication-efficiency-2}
\end{figure}

\begin{figure}[t]
    \centering
    \includegraphics[width=0.9\linewidth]{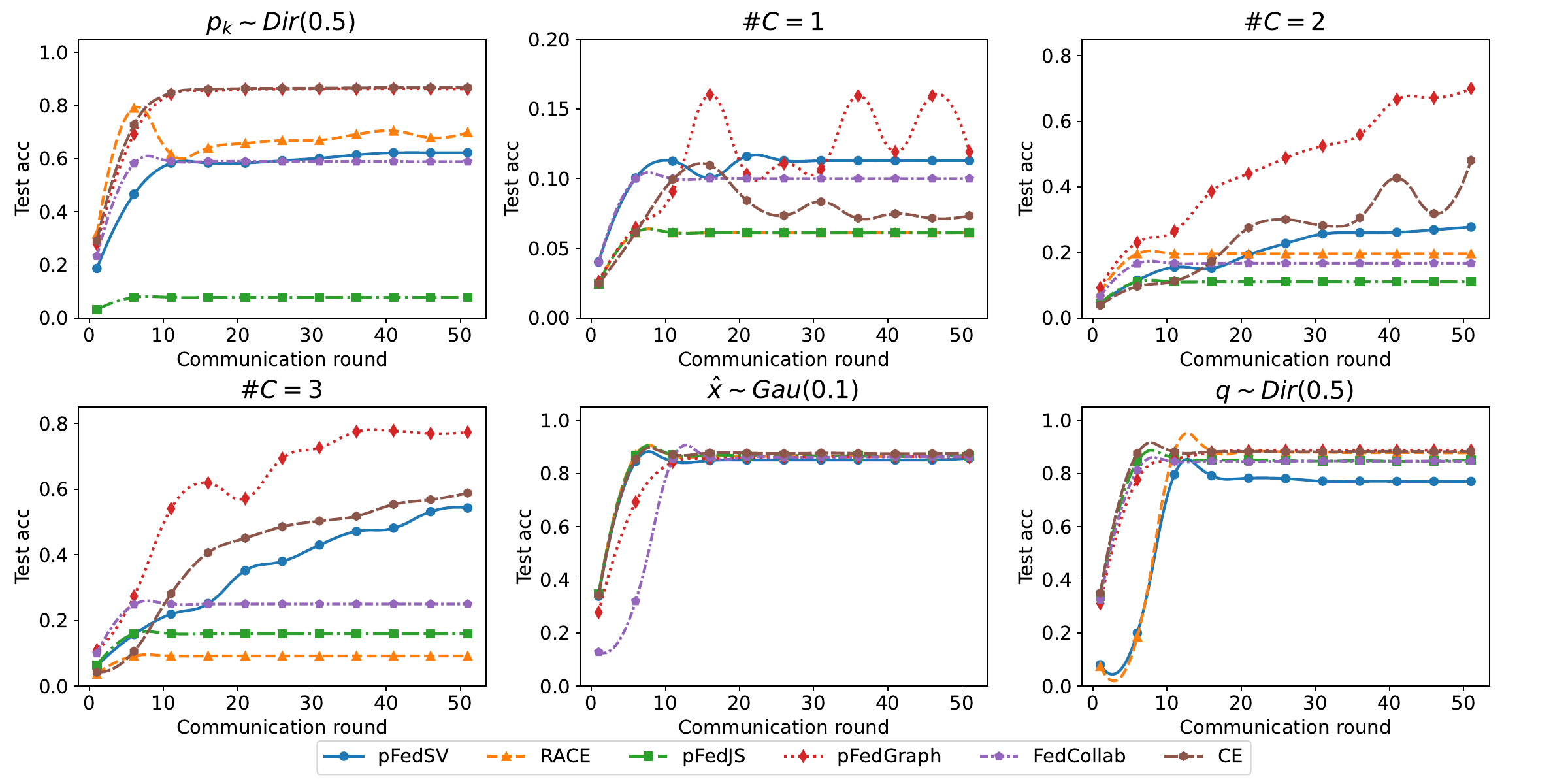}
    \caption{The training curves of different approaches on SVHN.}
    \label{fig:communication-efficiency-3}
\end{figure}

\begin{figure}[htp]
    \centering
    \includegraphics[width=0.9\linewidth]{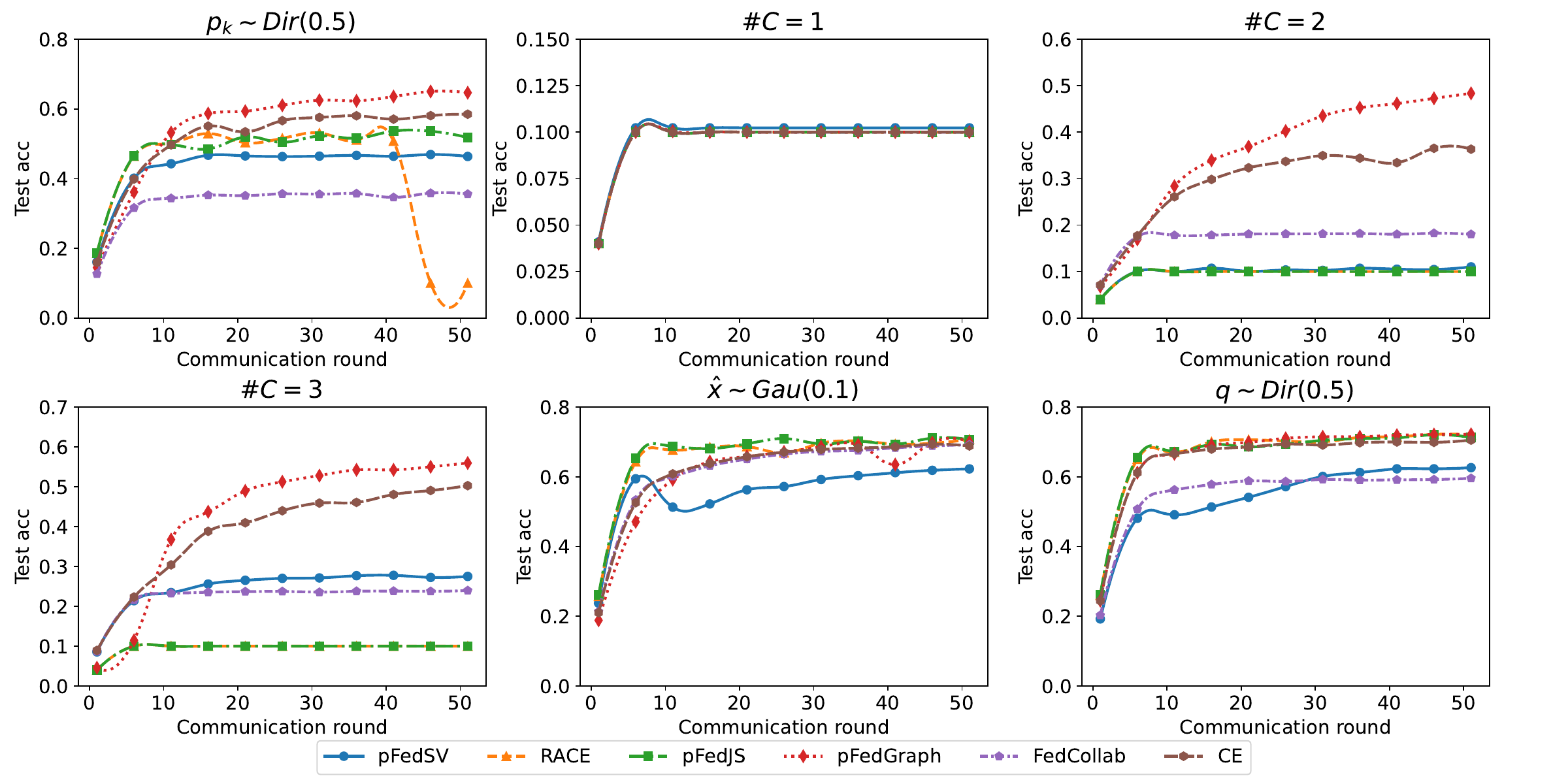}
    \caption{The training curves of different approaches on CIFAR-10.}
    \label{fig:communication-efficiency-4}
\end{figure}

\begin{figure}[htp]
    \centering
    \includegraphics[width=0.9\linewidth]{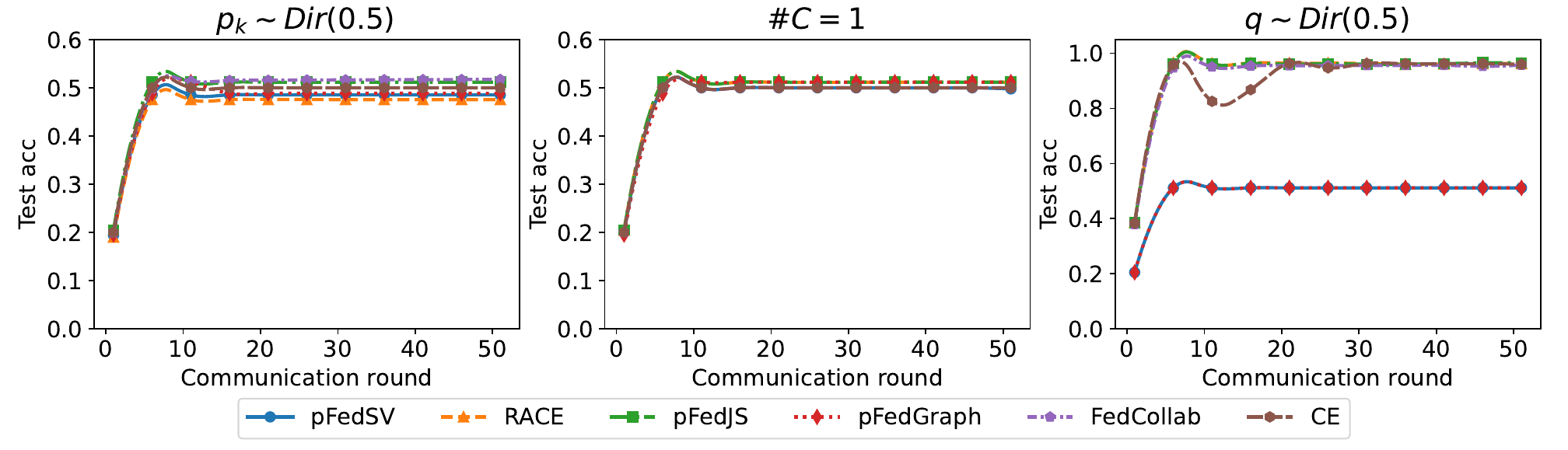}
    \caption{The training curves of different approaches on rcv1.}
    \label{fig:communication-efficiency-5}
\end{figure}

Figures \ref{fig:communication-efficiency-1}-\ref{fig:communication-efficiency-5} show the training curves of MNIST, FMNIST, SVHN, CIFAR-10, and rcv1 under different non-IID settings.
Under the non-IID setting $\mathbf{\hat{x}} \sim Gau(0.1)$, $p \sim Dir(0.5)$, and $p_{k} \sim Dir(0.5)$, training on these datasets is generally stable. However, when the number of classes \(\#C = 1, 2, 3\), the convergence and stability of different schemes vary significantly during the training period. For example, pFedGraph is relatively more unstable compared to other schemes when \(\#C = 1\). In addition, when \(\#C = 2, 3\), the accuracy of pFedSV, pFedGraph, and CE continues to improve across different datasets, indicating that these schemes require more communication rounds to achieve better performance under these conditions.

\subsection{Robustness to Local Updates}
\begin{figure}[htp]
    \centering
    \includegraphics[width=0.9\linewidth]{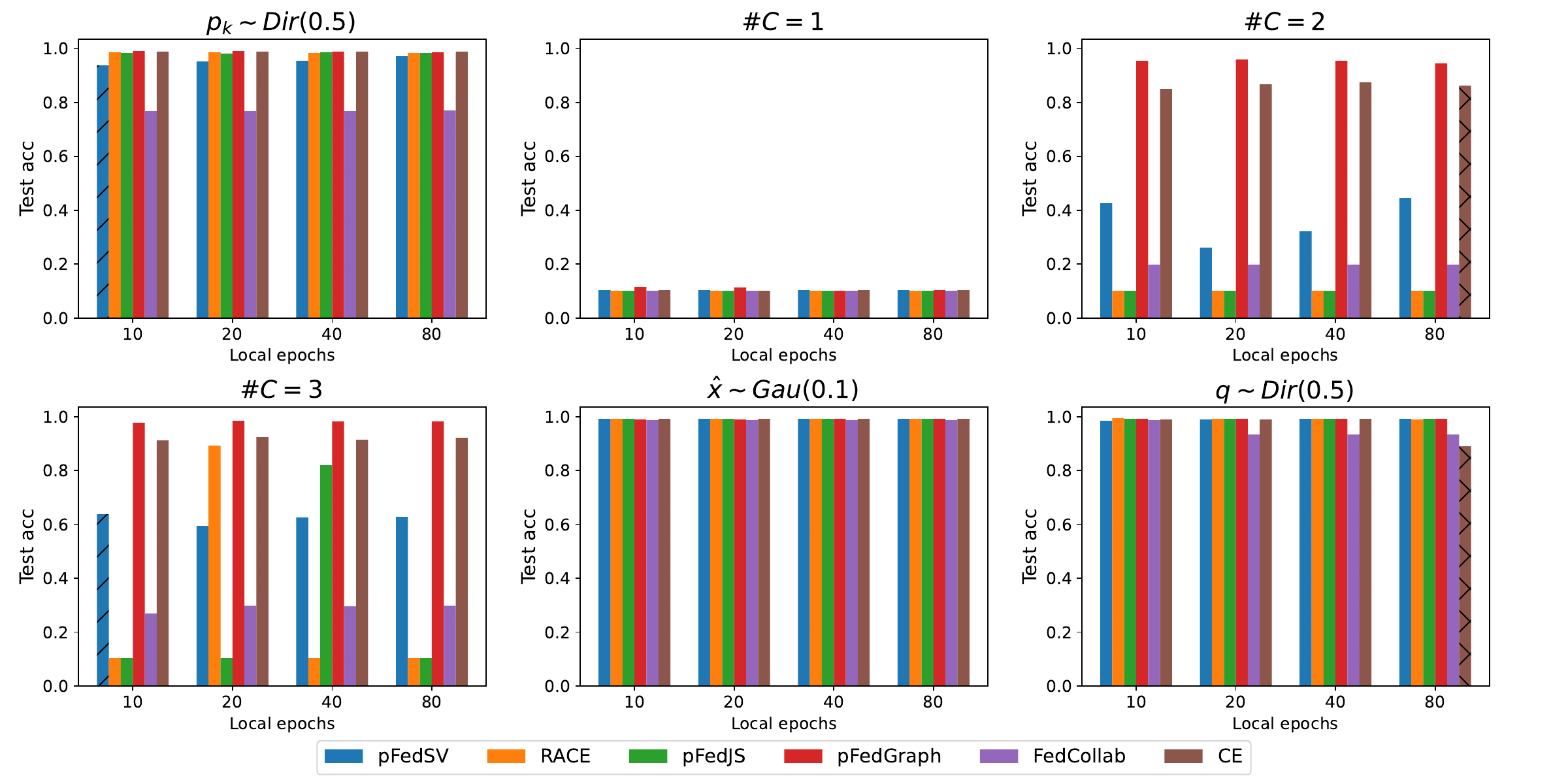}
    \caption{The test accuracy with different number of local epochs on MNIST.}
    \label{fig:local-epoch-1}
\end{figure}

\begin{figure}[htp]
    \centering
    \includegraphics[width=0.9\linewidth]{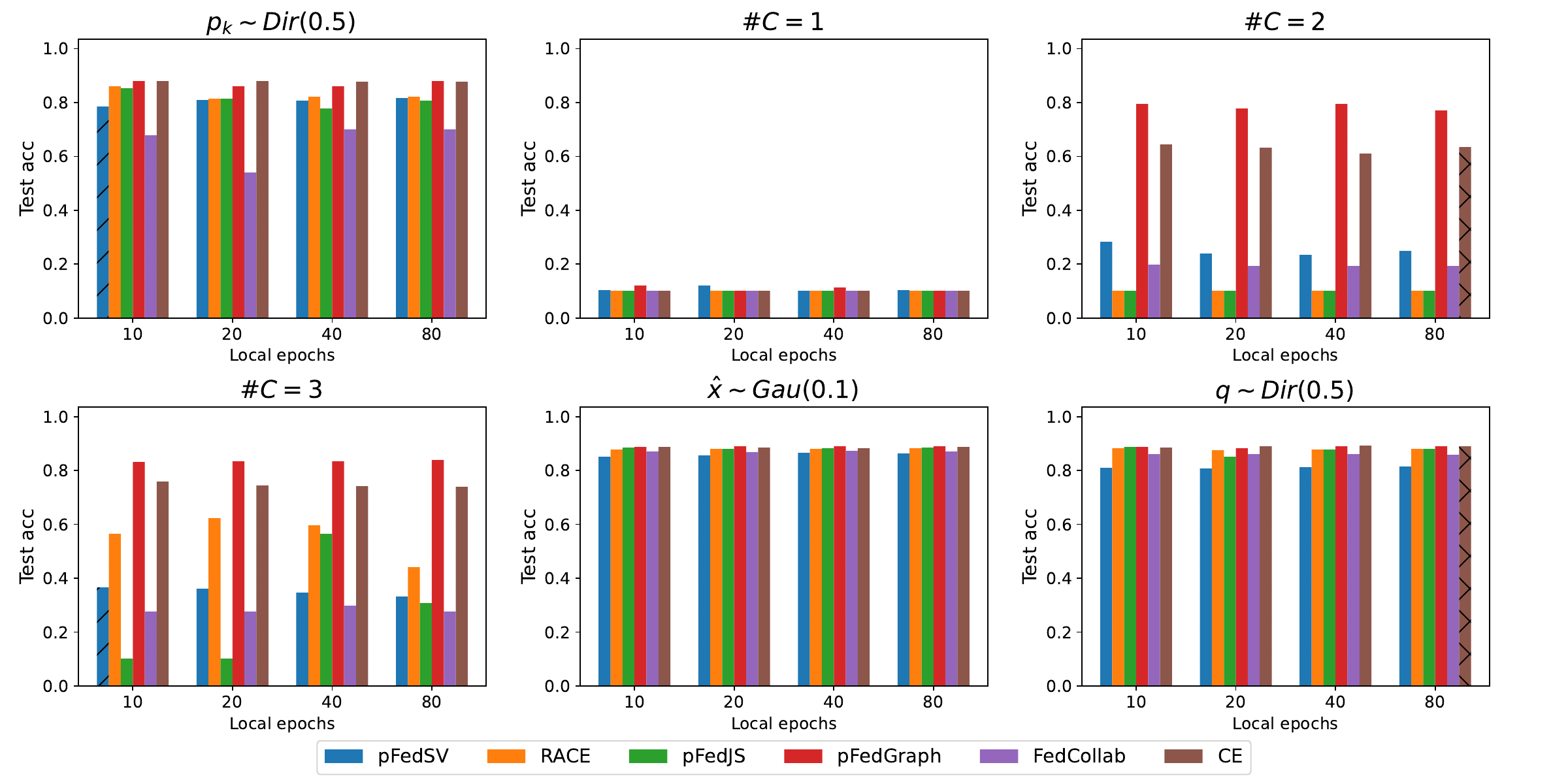}
    \caption{The test accuracy with different number of local epochs on FMNIST.}
    \label{fig:local-epoch-2}
\end{figure}

\begin{figure}[htp]
    \centering
    \includegraphics[width=0.9\linewidth]{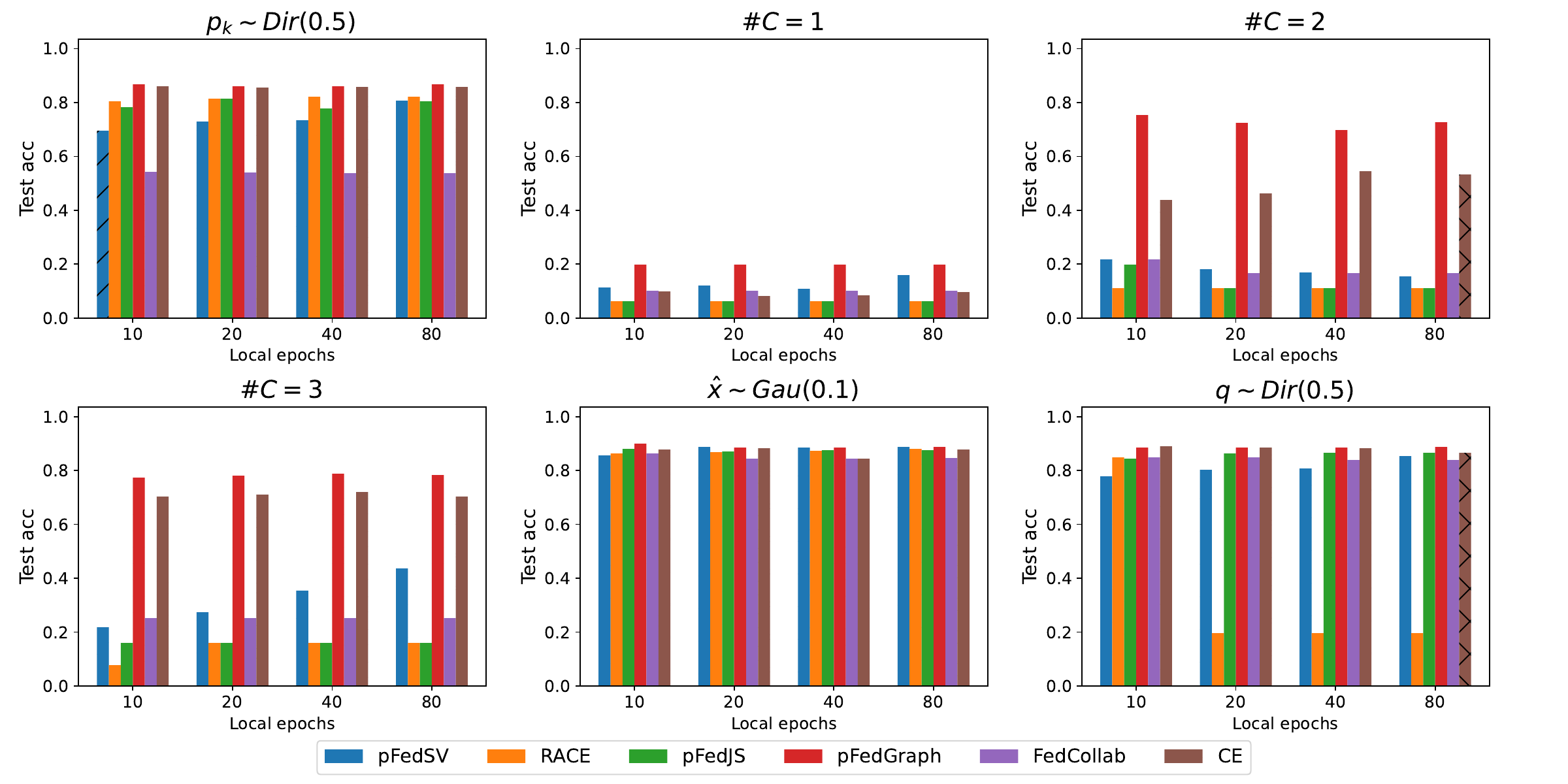}
    \caption{The test accuracy with different number of local epochs on SVHN.}
    \label{fig:local-epoch-3}
\end{figure}

Following \citep{NonIID-Benchmarking-ICDE-22}, we set the local epochs to 10, 20, 40, and 80 to test the robustness of the comparison schemes to local updates. As shown in Figures \ref{fig:local-epoch-1}-\ref{fig:local-epoch-3}, the number of local training epochs is crucial because it directly impacts the final accuracy. Contrary to intuition, a larger number of local epochs does not always lead to higher accuracy. When \(\#C=2, 3\), the accuracy of pFedJS and pFedSV on the test dataset initially increases or decreases in one direction, but as the number of local epochs continues to grow, the accuracy changes in the opposite direction. In the label distribution skew setting, FedCollab may experience overfitting as the number of epochs increases. The accuracy of CE occasionally decreases slightly in the quantity skew setting. RACE and pFedGraph perform very stably across various datasets and non-IID settings. Overall, these schemes achieve good performance with 10 local epochs. Increasing the number of local epochs further provides diminishing returns in terms of cost-effectiveness.

\subsection{Batch Size}
\begin{figure}[htp]
    \centering
    \includegraphics[width=0.9\linewidth]{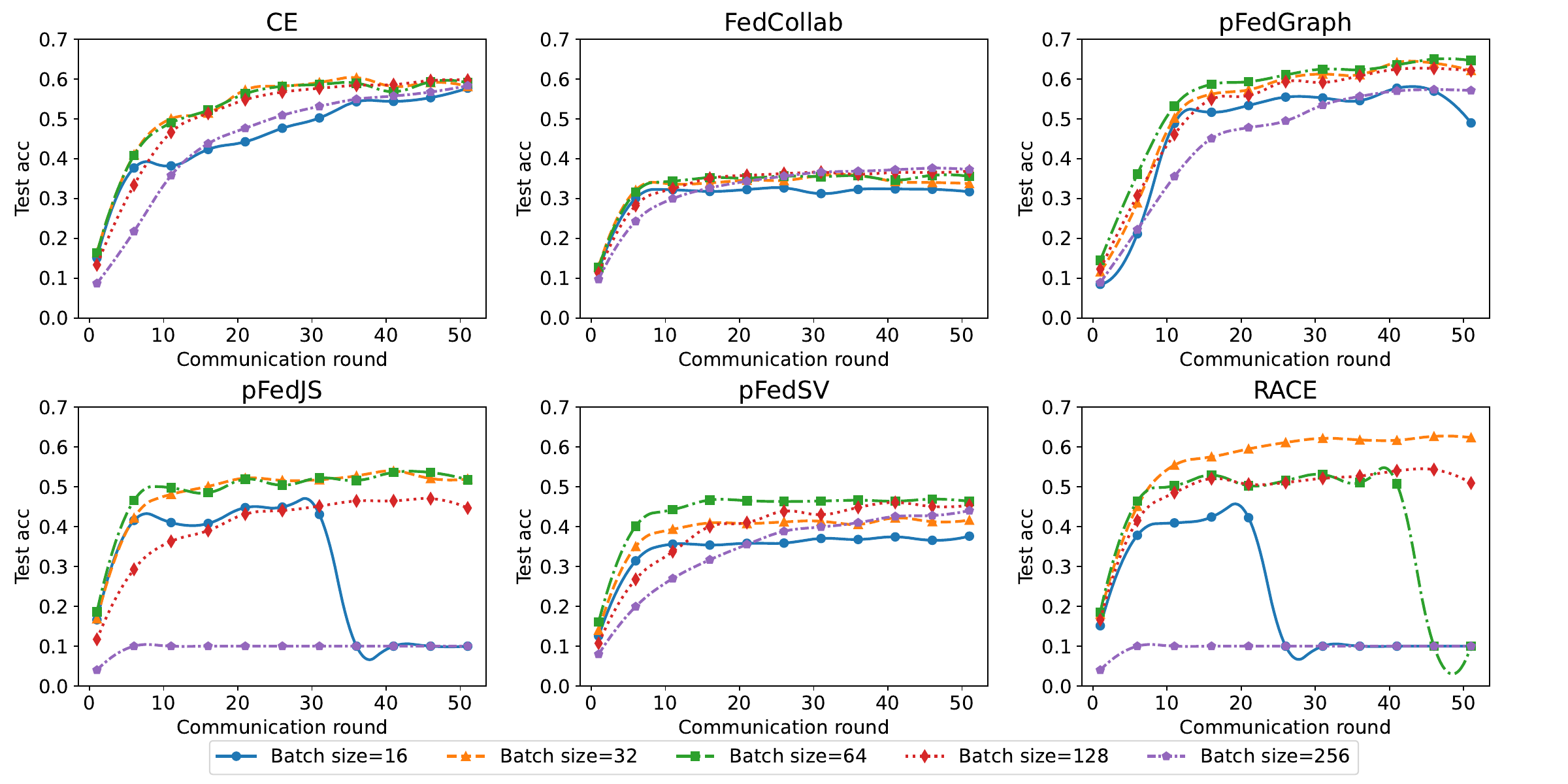}
    \caption{The training curves of different batch sizes on CIFAR-10 under $p_k \sim Dir(0.5)$ partition.}
    \label{fig:batchsize01}
\end{figure}

\begin{figure}[H]
    \centering
    \includegraphics[width=0.9\linewidth]{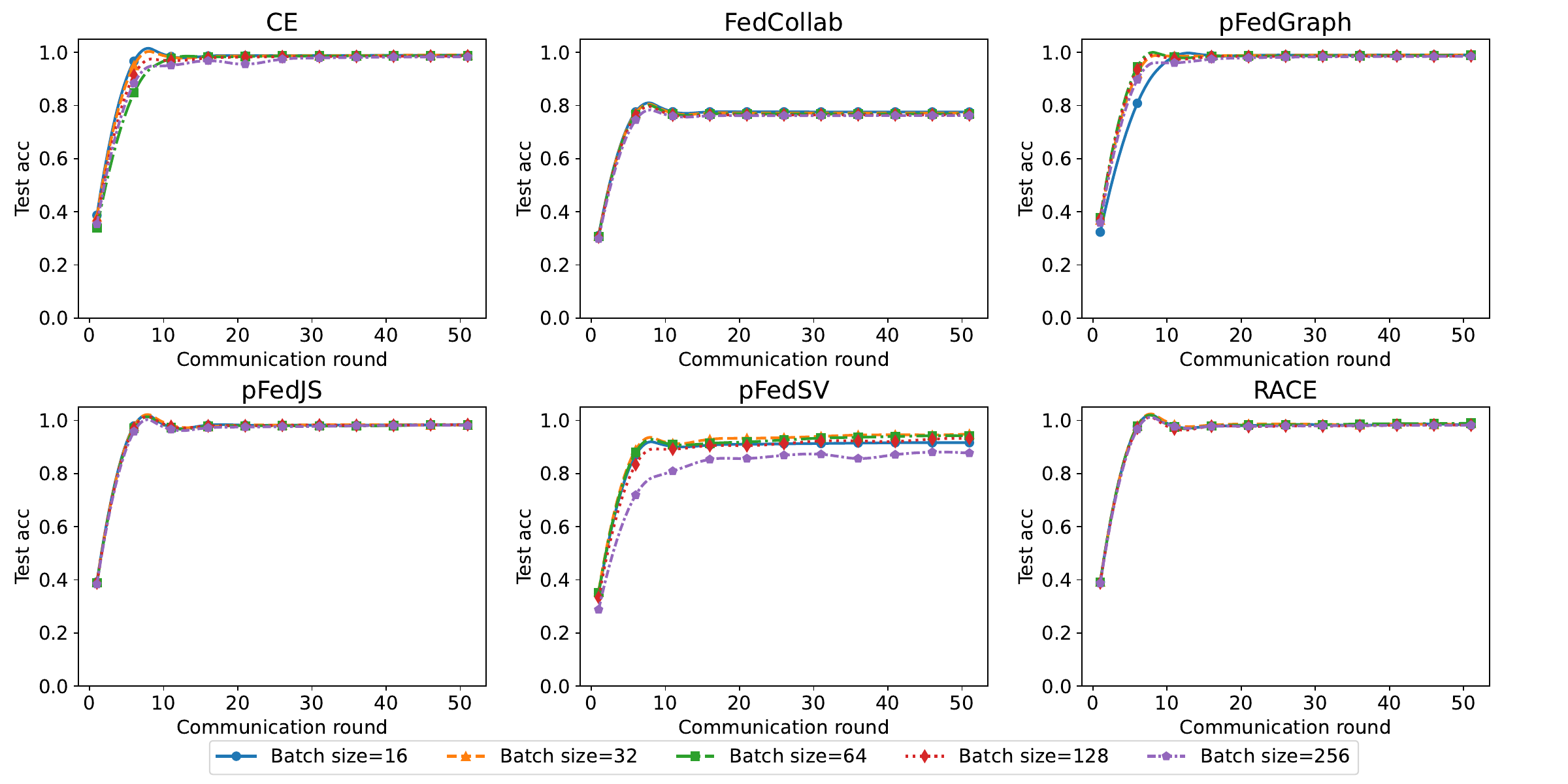}
    \caption{The training curves of different batch sizes on MNIST under $p_k \sim Dir(0.5)$ partition.}
    \label{fig:batchsize02}
\end{figure}

Figures \ref{fig:batchsize01}-\ref{fig:batchsize02} present the training curves for the six schemes with batch sizes ranging from 16 to 256. As expected, both excessively large and small batch sizes result in poor accuracy. Notably, in the pFedJS and RACE schemes of Figure \ref{fig:batchsize01}, the accuracy is particularly low when the batch size is 256, which may be due to getting stuck in local optima or having weak generalization ability. Additionally, when the batch size is 16, the accuracy drops sharply during training, likely due to the collaborative clients remaining largely unchanged and leading to overfitting while training personalized models. Therefore, selecting an appropriate batch size is crucial when training personalized models.

\newpage
\section*{NeurIPS Paper Checklist}

\begin{enumerate}

\item {\bf Claims}
    \item[] Question: Do the main claims made in the abstract and introduction accurately reflect the paper's contributions and scope?
    \item[] Answer: \answerYes{} 
    \item[] Justification: The main claims presented in the abstract and introduction of the paper align well with the contributions and scope detailed in the full manuscript. These sections effectively summarize the key points and core innovations of the study, providing a clear and concise overview that matches the in-depth discussions and findings elaborated upon in the subsequent sections.

\item {\bf Limitations}
    \item[] Question: Does the paper discuss the limitations of the work performed by the authors?
    \item[] Answer: \answerYes{} 
    \item[] Justification: The data heterogeneity we studied and compared was conducted in the context of personalized federated learning scenarios.

\item {\bf Theory Assumptions and Proofs}
    \item[] Question: For each theoretical result, does the paper provide the full set of assumptions and a complete (and correct) proof?
    \item[] Answer: \answerYes{} 
    \item[] Justification: Yes, after the previous assumptions, we put these six methods under the same framework for measurement and comparison.

    \item {\bf Experimental Result Reproducibility}
    \item[] Question: Does the paper fully disclose all the information needed to reproduce the main experimental results of the paper to the extent that it affects the main claims and/or conclusions of the paper (regardless of whether the code and data are provided or not)?
    \item[] Answer: \answerYes{} 
    \item[] Justification: Yes, we mentioned it in Experimental Studies and Experimental Setup Details. 

\item {\bf Open access to data and code}
    \item[] Question: Does the paper provide open access to the data and code, with sufficient instructions to faithfully reproduce the main experimental results, as described in supplemental material?
    \item[] Answer: \answerYes{} 
    \item[] Justification: Our experimental code is provided as URLs in the Abstract.

\item {\bf Experimental Setting/Details}
    \item[] Question: Does the paper specify all the training and test details (e.g., data splits, hyperparameters, how they were chosen, type of optimizer, etc.) necessary to understand the results?
    \item[] Answer: \answerYes{} 
    \item[] Justification:  Yes, we mentioned it in Experimental Studies and Experimental Setup Details. 

\item {\bf Experiment Statistical Significance}
    \item[] Question: Does the paper report error bars suitably and correctly defined or other appropriate information about the statistical significance of the experiments?
    \item[] Answer: \answerYes{} 
    \item[] Justification: We used error bars in Table \ref{tab:comparison}, specifically calculated using seeds 1, 2, and 3.

\item {\bf Experiments Compute Resources}
    \item[] Question: For each experiment, does the paper provide sufficient information on the computer resources (type of compute workers, memory, time of execution) needed to reproduce the experiments?
    \item[] Answer: \answerYes{} 
    \item[] Justification: We mentioned the computer resources used in Appendix \ref{append-exp-setup}.
    
\item {\bf Code Of Ethics}
    \item[] Question: Does the research conducted in the paper conform, in every respect, with the NeurIPS Code of Ethics \url{https://neurips.cc/public/EthicsGuidelines}?
    \item[] Answer: \answerYes{} 
    \item[] Justification: We have read the ethics review guidelines and have ensured that this paper fully conforms to them.

\item {\bf Broader Impacts}
    \item[] Question: Does the paper discuss both potential positive societal impacts and negative societal impacts of the work performed?
    \item[] Answer: \answerNo{} 
    \item[] Justification: This paper does not discuss any potential negative societal impacts of the research presented.
    
\item {\bf Safeguards}
    \item[] Question: Does the paper describe safeguards that have been put in place for responsible release of data or models that have a high risk for misuse (e.g., pretrained language models, image generators, or scraped datasets)?
    \item[] Answer: \answerNA{} 
    \item[] Justification: Our paper poses no such risks.

\item {\bf Licenses for existing assets}
    \item[] Question: Are the creators or original owners of assets (e.g., code, data, models), used in the paper, properly credited and are the license and terms of use explicitly mentioned and properly respected?
    \item[] Answer: \answerYes{} 
    \item[] Justification: Yes, in our work, we have duly cited the creators of the assets used. The referenced assets include works by creators \citet{Ding-NIPS-22, DSketch-NIPS-23, Shapley-TMC-24, FedCollab-ICML-23, Cosine-Similarity-ICML-23, cui2022collaboration} , whose contributions are acknowledged and referenced appropriately in accordance with academic standards and copyright practices.

\item {\bf New Assets}
    \item[] Question: Are new assets introduced in the paper well documented and is the documentation provided alongside the assets?
    \item[] Answer: \answerYes{} 
    \item[] Justification: We provide documentation and comments in the code.

\item {\bf Crowdsourcing and Research with Human Subjects}
    \item[] Question: For crowdsourcing experiments and research with human subjects, does the paper include the full text of instructions given to participants and screenshots, if applicable, as well as details about compensation (if any)? 
    \item[] Answer: \answerNA{} 
    \item[] Justification: There were no human subjects involved in this project.

\item {\bf Institutional Review Board (IRB) Approvals or Equivalent for Research with Human Subjects}
    \item[] Question: Does the paper describe potential risks incurred by study participants, whether such risks were disclosed to the subjects, and whether Institutional Review Board (IRB) approvals (or an equivalent approval/review based on the requirements of your country or institution) were obtained?
    \item[] Answer: \answerNA{} 
    \item[] Justification: We do not have any research with human subjects that forms a part of this work.

\end{enumerate}

\end{document}